\documentclass[sigconf]{acmart}
\renewcommand\footnotetextcopyrightpermission[1]{} 
\AtBeginDocument{%
  \providecommand\BibTeX{{%
    \normalfont B\kern-0.5em{\scshape i\kern-0.25em b}\kern-0.8em\TeX}}}

\setcopyright{acmcopyright}
\copyrightyear{2017}
\acmYear{2017}
\acmDOI{XXXXXXX.XXXXXXX}

\acmConference[MM '22]{Proceedings of the 30th ACM International Conference on Multimedia}{October 10--14, 2022}{Portugal, Lisbon}
 \acmBooktitle{Proceedings of the 30th ACM International Conference on Multimedia (MM '22), October 10--14, 2022, Portugal, Lisbon}

\acmISBN{978-X-XXXX-XXXX-X/18/06}
\usepackage{makecell}
\usepackage{multirow}
\usepackage{float}
\usepackage{threeparttable}
\usepackage{enumitem}
\usepackage{newfloat}
\usepackage{listings}
\usepackage{bbding}
\usepackage{fancyhdr}
\usepackage{xcolor}
\usepackage{tablefootnote}
\usepackage{algorithm}
\usepackage{algpseudocode}
\usepackage{stfloats}
\usepackage{bm}
\usepackage{amsmath}
\usepackage[switch]{lineno}

\begin{document}

\newcolumntype{L}[1]{>{\raggedright\arraybackslash}p{#1}}
\newcolumntype{C}[1]{>{\centering\arraybackslash}p{#1}}
\newcolumntype{R}[1]{>{\raggedleft\arraybackslash}p{#1}}
\title{Exploring User Retrieval Integration towards Large Language Models for Cross-Domain Sequential Recommendation}
\author{Tingjia Shen}
\email{jts_stj@mail.ustc.edu.cn}
\affiliation{%
  \institution{University of Science and Technology of China}
  \city{Hefei}
  \country{China}
}
\author{Hao Wang}
\email{wanghao3@ustc.edu.cn}
\affiliation{%
  \institution{University of Science and Technology of China}
  \city{Hefei}
  \country{China}
}
\author{Jiaqing Zhang}
\email{jiaqing.zhang@mail.ustc.edu.cn}
\affiliation{%
  \institution{University of Science and Technology of China}
  \city{Hefei}
  \country{China}
}
\author{Sirui Zhao}
\email{sirui@mail.ustc.edu.cn}
\affiliation{%
  \institution{University of Science and Technology of China}
  \city{Hefei}
  \country{China}
}
\author{Liangyue Li}
\email{liliangyue.lly@alibaba-inc.com}
\affiliation{%
  \institution{China Alibaba Group}
  \city{Hangzhou}
  \country{China}
}
\author{Zulong Chen}
\email{zulong.czl@alibaba-inc.com}
\affiliation{%
  \institution{China Alibaba Group}
  \city{Hangzhou}
  \country{China}
}
\author{Defu Lian}
\email{liandefu@ustc.edu.cn}
\affiliation{%
  \institution{University of Science and Technology of China}
  \city{Hefei}
  \country{China}
}
\author{Enhong Chen}
\email{cheneh@ustc.edu.cn}
\affiliation{%
  \institution{University of Science and Technology of China}
  \city{Hefei}
  \country{China}
}

\renewcommand{\shortauthors}{Anonymous Author, et al.}

\begin{abstract}

Cross-Domain Sequential Recommendation~(CDSR) aims to mine and transfer users’ sequential preferences across different domains to alleviate the long-standing cold-start issue.
Traditional CDSR models capture collaborative information through user and item modeling while overlooking valuable semantic information. Recently, Large Language Model~(LLM) has demonstrated powerful semantic reasoning capabilities, motivating us to introduce them to better capture semantic information.
However, introducing LLMs to CDSR is non-trivial due to two crucial issues: seamless information integration and domain-specific generation. To this end, we propose a novel framework named \textbf{URLLM}, which aims to improve the CDSR performance by exploring the \underline{\textbf{U}}ser \underline{\textbf{R}}etrieval approach and domain grounding on \underline{\textbf{LLM}} simultaneously.
Specifically, we first present a novel dual-graph sequential model to capture the diverse information, along with an alignment and contrastive learning method to facilitate domain knowledge transfer. Subsequently, a user retrieve-generation model is adopted to seamlessly integrate the structural information into LLM, fully harnessing its emergent inferencing ability. Furthermore, we propose a domain-specific strategy and a refinement module to prevent out-of-domain generation. Extensive experiments on Amazon demonstrated the information integration and domain-specific generation ability of URLLM in comparison to state-of-the-art baselines. Our code is available at \href{https://github.com/TingJShen/URLLM}{\textcolor{blue}{https://github.com/TingJShen/URLLM}}

\end{abstract}

\ccsdesc[500]{Information systems~Information retrieval}
\ccsdesc[500]{Computing methodologies~Natural language generation}
\keywords{Cross-Domain Sequential Recommendation, Large Language Model, Cold-Start Recommendation}

\maketitle

\section{Introduction}

Sequential Recommendation (SR), focused on suggesting the next item for a user based on their past sequential interactions to capture dynamic user preferences, has gained significant attention in commercial, social, and diverse scenarios~\cite{zhang2022clustering,DBLP:journals/corr/HidasiKBT15,wang2019mcne}. However, SR methods within a single domain usually encounter the long-standing cold-start issue~\cite{DBLP:conf/sigir/ScheinPUP02}, i.e., it is challenging to perform personalized recommendations for users with few interaction records.

To address the issue, Cross-Domain Sequential Recommendation~(CDSR) has garnered considerable attention in the field of recommendation systems, aiming to mine and transfer users’ sequential preferences across different domains~\cite{10.1145/3583780.3614657,yin2024learning}. Pioneer works like $\pi$-net~\cite{10.1145/3331184.3331200} and PSJNet~\cite{9647967} focused on designing knowledge transfer modules to capture cross-domain user preferences. Follow-up works like MIFN~\cite{DBLP:journals/tkdd/MaRCRZLMR22} and DA-GCN~\cite{ijcai2021p342} further borrowed the powerful strength of Graph Neural Networks~(GNNs) to model the high-order relationship across domains. These methods have been demonstrated to be effective for the CDSR problem. Despite the achieved results, most previous works overlook the valuable semantic information buried in item features~\cite{10.1007/978-981-16-5348-3_48,ijcai2021p342,zhang2024unified}, leading to skewed user preferences. 
Recently, the powerful emergent capabilities~\cite{DBLP:journals/corr/abs-2304-05332} of Large Language Models~(LLMs) have revolutionized the field of recommendation systems~\cite{DBLP:journals/corr/abs-2311-12338}, which can absorb item text features and inject pre-trained common knowledge into recommendation systems. Meanwhile, they can generate recommendations based on user preferences and historical data, enabling interactive and explainable recommendations~\cite{DBLP:journals/corr/abs-2307-02046}. LLMs also offer the flexibility to design tuning strategies for specific subtasks, such as determining whether to recommend an item or integrating collaborative information. This integration aligns the sequential behaviors of users with the language space, creating new modalities for recommendation~\cite{DBLP:conf/sigir/YuanYSLFYPN23, DBLP:journals/corr/abs-2311-10947}. Therefore, we are motivated by these encouraging capabilities to propose an LLM paradigm for the CDSR scenario.

\begin{figure}[t]
  \centering
  \includegraphics[width=0.95\linewidth]{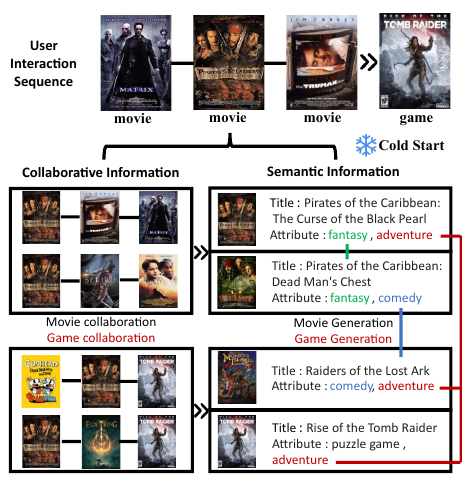}
  \caption{An illustration of cold-start CDSA task along with the various forms of information, the domain-specific demand on information and generation. The line linking attributes represent the structural-semantic information.}
  \label{fig:motivation}
\end{figure}

However, harnessing the explicit capabilities of LLMs in the CDSR scenario is non-trivial. Two essential issues arise:
(1) \textbf{Seamless Information Integration}: As shown in Figure~\ref{fig:motivation}, the CDSR task involves diverse formats of domain information, including collaborative and semantic information. Items exhibit intricate intrinsic structures respectively specified by the diverse information. To fully leverage the emergent capabilities of LLMs, it is crucial to integrate the structured information into LLMs in a seamless manner.
(2) \textbf{Domain-Specific Generation}. CDSR requires equipping the model with domain-specific information and constraining the generation process to remain confined within a specific domain. However, despite instructing the model to generate items within a particular domain, the uncontrollable nature of LLMs leads to 2\% to 20\% of generated content belonging to other domains. Figure~\ref{fig:motivation} illustrates an optimal pipeline for domain-specific generation, wherein distinct information aligns with specific generation processes.
Regarding issue (1), some efforts have been made to integrate collaborative and semantic information with LLMs. Pioneer works like UniSRec~\cite{10.1145/3534678.3539381} and CTRL~\cite{DBLP:journals/corr/abs-2306-02841} proposed to integrate semantic information through a discriminative LLM like BERT. As for collaborative information, BIGRec~\cite{DBLP:journals/corr/abs-2308-08434} adopts a post-processing approach, integrating through statistical results. RecInterpreter~\cite{DBLP:journals/corr/abs-2310-20487}, LLaRA~\cite{DBLP:journals/corr/abs-2312-02445}, and CoLLM~\cite{DBLP:journals/corr/abs-2310-19488} tend to capture collaborative information using an external traditional model and map it into the input token embedding space of LLM. However, these methods fail to seamlessly incorporate the emergent language reasoning capabilities of LLMs~\cite{li2023selfprompting,DBLP:conf/nips/Wei0SBIXCLZ22}, as evidenced by the misalignment between the pre-trained knowledge representation of LLMs and their embedding representations. This oversight is particularly evident in the absence of a seamless integration of collaborative information and structural-semantic information. Consequently, the integration of diverse information continues to pose a substantial challenge. 
In addressing issue (2), existing retrieval methodologies for LLMs, such as ICL~\cite{li2023selfprompting} and Agent-based LLM~\cite{zhang2023generative,huang2024understanding}, primarily yield common sense retrievals rather than domain-specific information. Furthermore, while the existing LLM-based recommendation model ~\cite{10.1145/3604915.3608857}~\cite{DBLP:journals/corr/abs-2308-08434} ~\cite{DBLP:journals/corr/abs-2310-14304} acknowledge the cross-domain capabilities of LLMs, they fail to impose constraints on the generation process of LLMs. This lack of restriction results in out-of-domain generations, significantly undermining the performance of CDSR. 

Towards these challenges, in this paper, we propose a novel framework named \textbf{URLLM}, addressing the cold-start issue by exploring a novel paradigm of \textbf{U}ser-\textbf{R}etrieval approach and domain grounding on \textbf{LLM} simultaneously. Firstly, we propose a dual graph sequence modeling model combining alignment and contrastive learning on an item-attribute graph and three item-item domain sequence graphs, aiming to model collaborative and structural-semantic information. Subsequently, we present a KNN user retriever to retrieve relevant user information for LLM. By leveraging LLM's extensive reasoning and few-shot learning capabilities in the integration of collaborative user information, we can achieve a seamless integration of diverse information. Finally, in order to preserve the domain-specific nature of both the input and generated responses in the LLM, we propose a domain differentiation strategy for user retrieval modules. Additionally, we introduce a refining mechanism to enhance the outputs of the LLM.

Extensive experiments on two datasets, including movie-game and art-office on Amazon following~\cite{10.1145/3534678.3539381} have demonstrated the effectiveness of the proposed framework. Moreover, through analysis of experimental results, we (1) recognize that the improvement brought by the types of information integrated into components of URLLM  positively correlated with the most crucial information in the dataset. (2) find there existing positive relation between the hit rate of retrieved users and the performance of the model. The main contributions could be summarized as follows:

\begin{itemize}[leftmargin=*]
    \item To our best knowledge, we are the first to study CDSR from a new perspective on the user retrieval paradigm with seamless information integration and domain-specific generation.

    \item 
    We develop a user retrieval bounded interaction paradigm between dual graph sequence modeling models and LLM. With the aid of the module, we can integrate structural-semantic and collaborative information into LLM in a seamless manner.
    \item 
    We introduce a domain differentiation strategy for user retrieval modules and a refinement module for the generated items of the LLM. The proposed module ensures that the integrated user information and generation are tailored to specific domains, aiming for domain-specific generation.
    
    \item Extensive experiments on two public datasets and ablation analysis validate that our URLLM framework unequivocally
affirms the information integration and domain-specific generation ability of our proposed framework.
\end{itemize}
\section{Related Work}
\subsection{Sequential Recommendation}
Sequential recommendation is a technique that aims to delve into and understand users' interest patterns by analyzing their historical interactions. Initially, techniques such as markov chain and matrix factorization were employed~\cite{10.1145/2911451.2911489}. However, with the emergence of neural networks, deep learning approaches like GRU4Rec~\cite{10.1145/3269206.3271761}, Hypersorec~\cite{wang2021hypersorec} and Caser~\cite{10.1145/3159652.3159656} were introduced to improve recommendation accuracy, while some efforts are made with clustering~\cite{han2023guesr}, denoising~\cite{han2024end4rec} or data regeneration~\cite{yin2024dataset}. Another notable technique in sequential recommendation is the attention mechanism. SASRec~\cite{8594844} and APGL4SR~\cite{yin2023apgl4sr}, for instance, utilize self-attention to independently learn the impact of each interaction on the target behavior. On the other hand, BERT4Rec~\cite{10.1145/3357384.3357895} incorporates bi-directional transformer layers after conducting pre-training tasks. In recent years, graph neural networks (GNNs) have gained attention for their ability to capture higher-order relationships among items. GCE-GNN~\cite{10.1145/3397271.3401142} constructs local session graphs and leverages information from other sessions to create a dense global graph for modeling the current session. SR-GNN~\cite{DBLP:journals/corr/abs-1811-00855} employs gated GNNs in session graphs to capture complex item transitions. To address the issue of data sparsity, contrastive mechanisms have been adopted in some works. CL4SRec~\cite{9835621} and CoSeRec~\cite{DBLP:journals/corr/abs-2108-06479} propose data augmentation approaches to construct contrastive tasks, which help alleviate the sparsity problem.

\subsection{Cross-Domain Sequential Recommendation}
Cross-Domain Sequential Recommendation~(CDSR) aims to improve recommendation performance for tasks involving items from different domains. Pioneering works in this field include $\pi$-Net~\cite{10.1145/3331184.3331200} and PSJNet~\cite{9647967}, which employ sophisticated gating mechanisms to transfer single-domain information. CD-SASRec~\cite{10.1007/978-981-16-5348-3_48} extends SASRec to the cross-domain setting by integrating the source-domain aggregated vector into the target-domain item embedding. DA-GCN~\cite{ijcai2021p342}, a GNN-based model, constructs a domain-aware graph to capture associations among items from different domains. Hybrid models that combine various techniques to capture item dependencies within sequences and complex associations between domains have also been proposed. RecGURU~\cite{10.1145/3488560.3498388} introduces adversarial learning to unify user representations from different domains into a generalized representation. UniSRec~\cite{10.1145/3534678.3539381} leverages item texts to learn more transferable representations for sequential recommendation. In comparison, C$^2$DSR~\cite{10.1145/3511808.3557262} employs a graphical and attentional encoder to capture item relationships. It utilizes two sequential objectives, in conjunction with a contrastive objective, to facilitate the joint learning of single-domain and cross-domain user representations, achieving significant progress.

\subsection{LLM-based Recommendation System}
Large Language Models~(LLMs) have been widely adopted as recommender systems to leverage item text features and enhance recommendation performance~\cite{10.1145/3523227.3546767, liu2023user,wu2023survey}. The majority of existing LLM-based recommenders operate in a tuning-free manner, utilizing pretrained knowledge to generate recommendations for the next item~\cite{sun-etal-2023-chatgpt, DBLP:journals/corr/abs-2308-14296}. For instance, CHAT-REC~\cite{DBLP:journals/corr/abs-2303-14524} employs ChatGPT to grasp user preferences and enhance interactive and explainable recommendations. GPT4Rec~\cite{DBLP:conf/ecom/LiZWXLM23} utilizes GPT-2 to generate hypothetical "search queries" based on a user's historical data, which are then queried using the BM25 search engine to retrieve recommended items. Another research direction in LLM-based recommendation focuses on designing tuning strategies for specific subtasks. TALLRec~\cite{10.1145/3604915.3608857}, for example, employs instruction-tuning to determine whether an item should be recommended. BIGRec~\cite{DBLP:journals/corr/abs-2308-08434} adopts a post-processing approach where recommendations are initially generated using LLM and then integrated with collaborative information through an ensemble method. RecInterpreter~\cite{DBLP:journals/corr/abs-2310-20487} and LLaRA~\cite{DBLP:journals/corr/abs-2312-02445} proposes a novel perspective by considering the "sequential behaviors of users" as a new modality for LLMs in recommendation, aligning it with the language space. CoLLM~\cite{DBLP:journals/corr/abs-2310-19488} captures collaborative information using an external traditional model and maps it into the input token embedding space of LLM, creating collaborative embeddings for LLM utilization.

Despite the significant progress achieved in the field of SR through traditional or LLM-based methods, these approaches tend to focus on a limited perspective of information formats. Consequently, there is an underutilization of information pertaining to the cold-start feature of CDSR, which is the primary focus of this paper. Moreover, although some previous works such as Tallrec~\cite{10.1145/3604915.3608857}, BIGRec~\cite{DBLP:journals/corr/abs-2308-08434}, and LLM-Rec~\cite{DBLP:journals/corr/abs-2310-14304} recognize the cross-domain capabilities of LLMs, there is currently a lack of an LLM-based structure specifically optimized for CDSR.

\section{PRELIMINARY}
\subsection{Problem Definition}
In this work, we consider a general CDSR scenario, where users in the user set $U=\{u_1,u_2,...,u_{|U|}\}$ have interactions with two product sets $X=\{x_1,x_2,...,x_{|X|}\}$ and $Y=\{y_1,y_2,...,y_{|Y|}\}$. Each user $u \in U$ has an interaction sequence $S_{u}=[i_1,i_2,..., i_t,...,i_k]$ representing the chronological order of items in two domains. The primary objective of CDSR is to train the model $M(X, Y, U, S)$ to predict the subsequent product $i_{k+1} \in {X\cup Y}$, where $S=\{S_1,S_2,...,S_{|U|}\}$ denotes the cross-domain interaction sequence.

\subsection{Instruct Tuning on LLMs}
In this paper, we use LLMs as the recommendation model $M$ for answer formulation. 
The LLMs should be fine-tuned to adeptly adapt to the data distribution and domain knowledge relevant to specific downstream tasks. The fine-tuning process involves meticulously crafting instruction data to guide the model's output scope and format, as detailed below:
\begin{equation}
max_{\Phi}\sum_{x_i, a_i\in T}\sum_{t=1}^{|a_i|}log(P_{\Phi}(a_{i,t}|x,a_{i,<t})),
\end{equation}
where $\Phi$ is the parameters of LLM to be optimized, $T$ is the training set, $a_{i,t}$ is the $t$-th token of the generated answer word, and $x$ is the input context which contains an instruction and a query question.

\section{Methodology}

\begin{figure*}[t]
  \centering
  \includegraphics[width=0.95\linewidth]{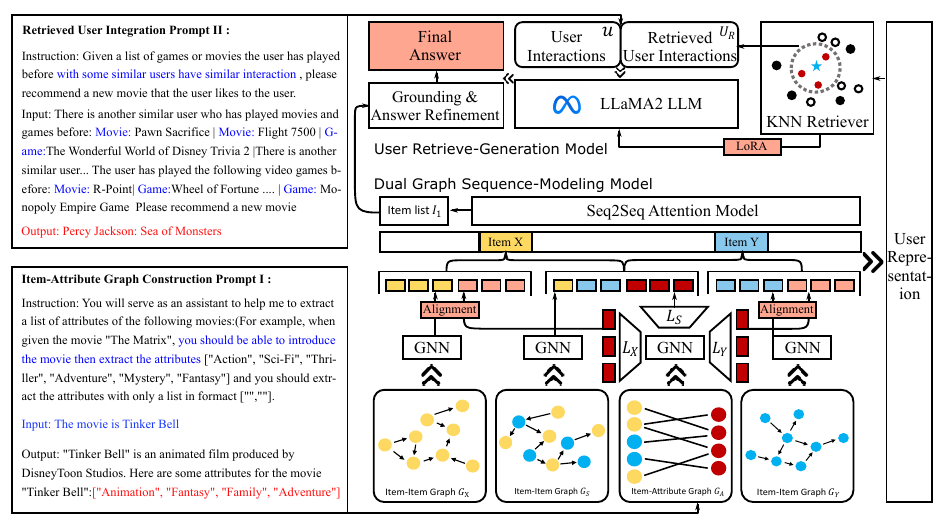}
  \caption{The overall framework of URLLM. The component on the left showcases exemplary prompts employed in graph construction and the similar user-augmented LLM module. On the right, the process is delineated, wherein the expansive reasoning and few-shot analogy capabilities of the LLM are harnessed, concomitantly integrating structured knowledge.}
  \label{fig:URLLM_structure}
\end{figure*}

To harness the inferential capabilities of Large Language Models~(LLMs) for blending LLMs with traditional models, we present URLLM illustrated in Figure~\ref{fig:URLLM_structure}.
Initially, we introduce a \textbf{Dual Graph Sequence-Modeling Model}. This model employs an LLM-enhanced item-attribute graph and an item-item sequence graph to encapsulate collaborative and structural-semantic information. This type of information is challenging to model with an isolated LLM. Then, we adopt a \textbf{User Retrieve-Generation Model} to retrieve the most similar users from the target domain and fuse structured text with collaborative information seamlessly into LLM. Finally, to make the output of our model directly match the real item in the correct domain, we combine BM25~\cite{10.1561/1500000019} and the Dual Graph Sequence-Modeling Model to refine the generated prediction.

\subsection{Dual Graph Sequence Modeling Model}
Aiming to model collaborative and semantic information, the construction of this model contains a graph construction module, a graph alignment module, and a contrastive self-attention module. 
\subsubsection{Graph Construction Module}
In order to capture valuable collaborative and semantic information, including their structural relationships, which can be effectively represented as a graph, we construct a graph based on rules due to the absence of such data in our dataset. The graph construction process comprises two modules: item-attribute graph construction and item-item graph construction. These modules establish collaborative relationships among items and structural-semantic relationships, respectively. For item-attribute graph construction, the Chain-of-Thought (COT) method~\cite{DBLP:conf/nips/Wei0SBIXCLZ22} is employed to utilize an LLM for item description. Subsequently, the model summarizes this description to generate output attributes for the product. For instance, in Figure~\ref{fig:URLLM_structure}, for the product I=Tinker Bell, the model provides an introduction denoted as $LLM(I)=Intro$ and generates the attribute reusing $I$ as $A_{I}=LLM(Intro, I)$. The undirected attribute graph, denoted as $G_A$, is constructed by linking items with attributes formulated below:
\begin{equation}   
G_A=(V,E), V=X\cup Y\cup A , E=\{(v,t)|v\in X\cup Y, t\in A\}.
\end{equation} 
where $V$ and $E$ denotes the vector and edges of graph $G_A$.

For item-item graph construction, we construct $G_S=G_{X+Y}$, $G_X$ and $G_Y$ respectively inspired by~\cite{DBLP:conf/cikm/CaoCSLW22}. Firstly, given user interaction sequence$S_{u}=[i_1,...,i_k]$, we split the interaction sequence into $S_{u}=S_{u}, S^X_{u}=S_{u}\cap X, S^Y_{u}=S_{u}\cap Y$. We then construct the directed item-item graph $G_S, G_{X}, G_{Y}$ by linking items before and after using the formula below:
\begin{equation}   
G_S=(V,E), V=X\cup Y, E=\{(v,t)|\exists j, v=i_{j}, t=i_{j+1}\},
\end{equation} 
\begin{equation}   
G_{X}=(V,E), V=X, E=\{(v,t)|\exists j, v=i^X_{j}, t=i^X_{j+1}\},
\end{equation} 
\begin{equation}   
G_{Y}=(V,E), V=Y, E=\{(v,t)|\exists j, v=i^Y_{j}, t=i^Y_{j+1}\},
\end{equation} 
where $V$ denotes the vector set, $E$ denotes the edge set, $i^X$ and $i^Y$ denotes item in $S^X_{u}$ and $S^Y_{u}$.

\subsubsection{Graph Alignment Module}

Recognizing the deficiency of differential knowledge across distinct domains in the existing item-attribute graph, it becomes imperative to investigate the acquisition of domain transfer information to enhance the item-attribute graphs. Domain transfer, in turn, enables the extraction of intricate intrinsic structures that are specifically specified by diverse information sources. However, the division of the item-attribute graph alone does not provide distinct information. Therefore, in this section, we propose a Graph Neural Network~(GNN) that incorporates an alignment loss function to align and integrate these fragmented pieces of information.

Given four graphs $G_A, G_S, G_{X}, G_{Y}$, We first initialize the graph embedding with $E_A^0\in \mathbb{R}^{|E_A^0|\times d_a}, E_S^0\in \mathbb{R}^{|E_S^0|\times d}, E^0_{X}\in \mathbb{R}^{|E^0_X|\times d}, E^0_{Y}\in \mathbb{R}^{|E^0_Y|\times d}$. Then, given these adjacency matrix $A_A, A_S, A_{X}, A_Y$, we construct $l$-layer GNN with output denoted as $E^{l}$ as below:
\begin{equation}   
\begin{aligned}
&E^{i}=Norm(A)E^{i-1}, i\in [1,l],
\end{aligned}
\end{equation} 
where $Norm(\cdot)$ represent the row-normalized function, $E^{i-1}$ represents the current graph convolutional layer and $E^i$ represents the next layer. Then, to fully encapsulate the graphical information across layers, we average the graph embeddings $E$ gaining $\hat{G}$ that yield item representations as:
\begin{equation}   
\begin{aligned}
&\hat{G}=\frac{1}{l}\Sigma_{i=1}^l(E^i)+E^0.
\end{aligned}
\end{equation} 
 We apply this procedure to $G_A, G_S, G_{X}, G_{Y}$ separately gaining $\hat{G_A},\\ \hat{G_S}, \hat{G_{X}}, \hat{G_{Y}}$, each with its corresponding adjacency matrix $A$ and initialized graph embedding $E^0$.
Then, To resolve differences in input and output dimensions in alignment,
we design linear projection models $L, L_X, L_Y$ with $d_a$ as input dim and $d$ as output dim to reform the hidden representation. 
Finally, the alignment loss function is designed below to transfer item-attribute graph knowledge of domains $X$ and $Y$: 
\begin{equation}   
\mathcal{L}_{al}=\Sigma_{S_X}||L_X(\hat{G_A})-\hat{G_{X}}||+\Sigma_{S_Y}||L_Y(\hat{G_A})-\hat{G_{Y}}||.
\end{equation} 
We then concat them together as the final representation $R$ of the item, gaining $R_S=Concat(L(\hat{G_A}), \hat{G_S}), R_X=Concat(L_X(\hat{G_A}), \hat{G_{X}}), \\R_Y=Concat(L_Y(\hat{G_A}), \hat{G_{Y}})$. With the alignment above the item-attribute and item-item graphs are modelized symmetrical and are successfully aligned to specific domains.
\subsubsection{Self-attention Contrastive Sequential Module}

\begin{figure}[t]
  \centering
  \includegraphics[width=0.95\linewidth]{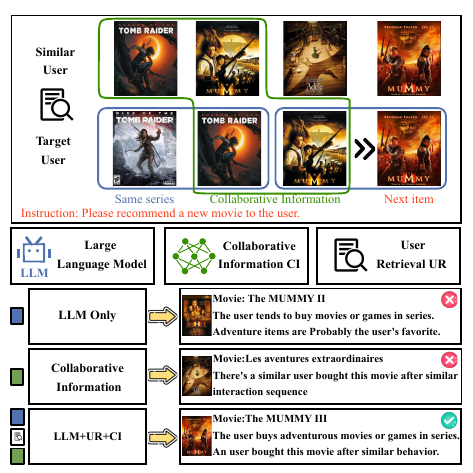}
  \caption{The example case illustrates the importance of LLM inferencing and similar user retrieval.}
  \label{fig:cases}
\end{figure}
Having successfully obtained dual graph item representations from the aforementioned models, our objective is to capture users' sequential preferences based on their interaction sequences. This endeavor is crucial to promote the retrieval of analogous user behaviors, thereby ensuring the optimal utilization of the few-shot capabilities of LLMs. Similar to SASRec~\cite{DBLP:conf/icdm/KangM18}, we employ a multi-head self-attention layer and point-wise feed-forward layer to distinctly capture user preferences and enhance retrieval precision. The formal definition of sequence modeling is provided below:

\begin{equation}  
\begin{aligned} 
&H_S=Attention_S(S_u,R_S), H_X=Attention_X(S_u^X,R_{X}), \\ &H_Y=Attention_Y(S_u^Y,R_{Y}),
\end{aligned}
\end{equation}

Denoting user preferences in the full domain, domain X, and domain Y as $H_S, H_X, H_Y$ respectively, the training target of the sequential module focuses on predicting the next model. We incorporate a discriminator $D_X,D_Y$ with a softmax function to assign scores to each item and then maximize the score for the ground-truth answer. The training loss is defined as follows:
\begin{equation}   
\mathcal{L}_t=
\left\{
     \begin{array}{lr}
     -log(P(i_k|D_X(H_S+H_X))) &i_k\in X  \\
     -log(P(i_k|D_Y(H_S+H_Y))) &i_k\in Y.  
     \end{array}
\right.
\end{equation} 
where $i_k$ denotes the item to be recommended.

To handle the negative transfer challenge on Cross-Domain Sequential Recommendation~(CDSR), we adopt sequence corruption and random noise integration to gain negative sequence samples. The formalized definition is demonstrated below:
\begin{equation}  
\begin{aligned}   
&\hat{H_X}=Attention_X(\hat{S_X},R_{X})+\Delta, \hat{H_Y}=Attention_B(\hat{S_Y},R_{Y})+\Delta,
\end{aligned}   
\end{equation}
where $\hat{S_X}$ and $\hat{S_Y}$ randomly replaces X and Y domain items. $\Delta$ is a random noice with $\left\|\Delta\right\|_{2}=\epsilon $ and $\Delta=\bar{\Delta}\odot sign(\mathbf{e}_i),\bar{\Delta}\in\mathbb{R}^d(0,1)$. The contrastive training loss is denoted as below:
\begin{equation} 
\begin{aligned}   
\mathcal{L}_{ct}&=\Sigma(-log(D_X(H_S+H_X))+(1-D_X(H_S+\hat{H_Y})))\\
&+\Sigma(-log(D_Y(H_S+H_Y))+(1-D_Y(H_S+\hat{H_X})))
\end{aligned}.
\end{equation} 
to maximize the infomax objective between domain preferences A and B. The final optimization loss is as follows:
\begin{equation} 
\mathcal{L}=\mathcal{L}_t+\lambda\mathcal{L}_{ct}+\gamma\mathcal{L}_{al}.
\end{equation}
By combining contrastive alignment methodology for dual graph inherent knowledge preferences $H_S, H_A, H_B$, which is hard for LLM to model. Furthermore, the dual graph sequence-modeling model is also a recommendation model with answer $I_1=[i_1,...,i_{|I|}]$, and the answer can be utilized to refine LLM's generation afterward.
\subsection{User Retrieve-Generation Model}

In the preceding section, we obtained high-quality structured user preferences. However, LLM cannot directly leverage this representation information. In order to fully exploit the emergent capabilities of LLMs, such as few-shot learning and language inferencing, we need a model that translates the representation information into textual information, preferably in the form of similar example data. Therefore, the following retrieve-augmented-generation contains a KNN retriever, a LoRA-tuned LLM augmented by the target user and retrieved user interactions, and an answer refinement structure.
\subsubsection{KNN retriver}

\begin{algorithm}[b]
\renewcommand{\algorithmicrequire}{\textbf{Input:}}
\renewcommand{\algorithmicensure}{\textbf{Output:}}
\caption{Overall pseudo code of URLLM}
\begin{algorithmic}[1]
\Require Product in two domain $X$ and $Y$, suppose $X_{id}<Y_{id}$. Training user set $U$, user interaction sequence $S$, user $u$
\Ensure User's next possible item list $I$
\State Train $M(S_{u}, X, Y)=E'_U,E'_u,I_1$ \ \ \ \ \ \ \ \ \ \ \ \ \ \ \Comment{Prompt I is used here}
\State Retrieve $U_r=KNN(u,U,E)$, $U_t=KNN(U,U,E)$
\State Train $LLM$ with $U_t$\ \ \ \ \ \ \ \ \ \ \ \ \ \ \Comment{Prompt II is used here}
\State $I_2=LLM(U_r,S_u)$
\State $maxI=max\{I_2[0:m]\}, minI=min\{I_2[0:m]\}$
\State $I=I_2$
\If{$maxI>X_{id}$ in recommend domain A}
    \State $I=I_1$
\EndIf
\If{$minI>Y_{id}$ in recommend domain B}
    \State $I=I_1$
\EndIf
\end{algorithmic}
\label{alg:URLLM}
\end{algorithm}

  We employ a KNN retrieval model to query the training users using $KNN(u)$ to retrieve its k-nearest neighbors, denoted as N, based on a distance function $d(\cdot,\cdot)$, specifically using the inner product, as they are already normalized beforehand. The formal retrieval procedure is presented below:
\begin{equation}
\begin{aligned}
&E'_X=H_S+H_X, E'Y=H_S+H_Y\\
&U_r=KNN(u,U,E)\\
&=
\left\{
     \begin{array}{lr}
Top_k(min(d({E'_X}_u,{E'_X}_{u_t}), u_t\in U))&domain\ A  \\
Top_k(min(d({E'_Y}_u,{E'_Y}_{u_t}), u_t\in U))&domain\ B, 
    \end{array}
\right.
\end{aligned}
\end{equation}
where $U$ denotes the user in the training set, $U_r$ denotes the retrieved users. 
\subsubsection{User Retrieval Augmented LLM}
After gaining high-quality similar users, we then integrate this high-quality knowledge using few-shot learning, proven to be one of the most effective ways to enhance the emergent ability of LLM. Specifically, we form the prompt as `There are similar users who bought these items:' with $S_{U_r}=\{S_{u_1},..., S_{u_k}\},u_1,...,u_k\in U_r$, and `The user has bought these items:' with $S_u$. The actual prompt is illustrated in prompt II, Figure~\ref{fig:URLLM_structure}.
However, there exists a distribution shift between original instruction tuning on the CDSR task and similar user integration CDSR task. We adopt the same KNN retrieval model on training users but choose from top-2 instead of top-1 (top-1 is the same user) to construct training prompts. Suppose the prompt is denoted as $P$, the optimization of user-integration tuning loss is as follows:
\begin{equation}
\mathcal{L}_r=max_{\Phi_L}\sum_{p\in P}\sum_{t=1}^{|p|}log(P_{\Phi+\Phi_L}(p_t|p_{<t})).
\end{equation}
$\Phi_L$ is the LoRA parameters and we only update LoRA parameters during the training process. However, due to the uncontrollability of LLM outputs, even when prompting and training explicitly for the generation of products in domain $X$, there is still about 2\% to 20\% output of products from domain $Y$ during actual inference. Therefore, a subsequent answer refinement module is required to further optimize the outputs of the LLM.

\subsubsection{Answer Refinement Structure}
We first adopt a BM25 retrieval model to ground the space of recommendation language space to actual item space with $I_2$. Subsequently, to stably identify whether LLM has influenced out-of-domain items, we consider the top-m grounding items. If one of the answers runs out of domain, we will consider adopting $I_1$ from the dual graph sequence-modeling model. In our experiment, we set m=5 to gain an answer stably. The overall algorithm is delineated in algorithm~\ref{alg:URLLM}.

Finally, by combining a dual graph sequence-modeling model and a controllable answer selection model, we leverage the expansive reasoning and few-shot analogy capabilities of the LLM on integrating collaborative user information. This approach not only addresses the cold-start problem but also tackles the challenges of aligning cross-domain information in collaborative recommendation systems. To provide a clearer illustration of how different components of URLLM contribute to its performance, we present example cases in Figure ~\ref{fig:cases}. 
\section{Experimental Settings}

\subsection{Datasets} 
To demonstrate the performance of our proposed model, we conduct experiments on two publicly available datasets from the Amazon platform~\cite{jin2023amazon}. This forms two distinct cross-domain scenarios: \textbf{Movie-Game} and \textbf{Art-Office}. The Movie-Game dataset, preprocessed by~\cite{bao2023bistep}, is sparser and contains many cold-start users. The Art-Office dataset exhibits a more standardized distribution and is preprocessed by~\cite{DBLP:conf/kdd/HouMZLDW22}. We further refine the latter by filtering users with fewer than 3 item interactions. Detailed descriptions and statistics for both datasets\footnote{The processed dataset will be open once accepted.} are provided in Table~\ref{tab:Datasetdescribe}.

 \begin{table}[t]
    \centering
    \caption{The detailed description and statistics of datasets.}
    \label{tab:Datasetdescribe}
    \begin{tabular}{cccccc}
    \toprule
        \textbf{Scenarios} & \textbf{\#Items}&\textbf{\#Train} &  \textbf{\#Valid} &  \textbf{\#Test}&  \textbf{Avg.length}  \\
    \midrule 
        Movie &71067       &\multirow{2}{*}{\makecell{35941}} &\multirow{2}{*}{\makecell{1775}} &\multirow{2}{*}{\makecell{3601}}&4.095\\
        Game&112233  &&&     &3.277\\
    \midrule
        Art &18639       &\multirow{2}{*}{\makecell{16000}} &\multirow{2}{*}{\makecell{1154}} &\multirow{2}{*}{\makecell{2000}}&6.386\\
        Office&19757  &&&     &8.263\\
    \bottomrule
    \end{tabular}
\end{table}

\subsection{Baselines}

\begin{table*}[t]
\centering
\caption{The overall performance of all baselines on the Movie-Game dataset.}
\label{tab:GM_datasets_benchmark}
\begin{tabular}{ccccccccccc}
\toprule
Baseline Type&Method & HR@1 & HR@5 & HR@10 & HR@20 & MRR & NG@1 & NG@5 & NG@10 & NG@20 \\
\midrule
\multirow{4}{*}{\makecell{Traditional\\ Baselines}}
&LightGCN & 0.0033 & 0.0064 & 0.0099 & 0.0147 & 0.0054 & 0.0033 & 0.0063 & 0.0087 & 0.0105 \\
&SASRec& 0.0013 & 0.0038 & 0.0049 & 0.0075 & 0.0029 & 0.0013 & 0.0025 & 0.0029 & 0.0035 \\
&CoSeRec & 0.0036 & 0.0112 & 0.0174 & 0.0251 & 0.0078 & 0.0036 & 0.0081 & 0.0090 & 0.0116 \\
&$C^2DSR$ & 0.0047 & 0.0124 & 0.0181 & 0.0268 & 0.0089 & 0.0047 & 0.0085 & 0.0104 & 0.0125 \\
\midrule
\multirow{4}{*}{\makecell{LLM-based \\Baselines}}
&UniSRec & 0.0090 & 0.0220 & 0.0259 & 0.0287 & 0.0148 & 0.0090 & 0.0160 & 0.0173 & 0.0181 \\&BIGRec & 0.0067 & 0.0194 & 0.0305 & 0.0444 & 0.0168 & 0.0067 & 0.0148 & 0.0183 & 0.0233 \\
&GPT4Rec& 0.0088 & 0.0322 & 0.0410 & 0.0515 & 0.0198 & 0.0088 & 0.0208 & 0.0234 & 0.0266 \\
&CoLLM & 0.0101 & 0.0202 & 0.0354 & 0.0455 & 0.0171 & 0.0101 & 0.0146 & 0.0193 & 0.0219 \\
\midrule
Our Work&URLLM & \textbf{0.0105} & \textbf{0.0333} & \textbf{0.0416} & \textbf{0.0522} & \textbf{0.0211} & \textbf{0.0105} & \textbf{0.0221} & \textbf{0.0248} & \textbf{0.0298} \\
\bottomrule
\end{tabular}
\end{table*}
\begin{table*}[t]

\centering
\caption{The overall performance of all baselines on the Art-Office dataset.}
\label{tab:AO_datasets_benchmark}
\begin{tabular}{ccccccccccc}
\toprule
Baseline Type&Method & HR@1 & HR@5 & HR@10 & HR@20 & MRR & NG@1 & NG@5 & NG@10 & NG@20 \\
\midrule
\multirow{4}{*}{\makecell{Traditional\\ Baselines}}
&LightGCN & 0.0105 & 0.0225 & 0.0255 & 0.0325 & 0.0198 & 0.0104 & 0.0133 & 0.0163 & 0.0184 \\
&SASRec & 0.0055 & 0.0120 & 0.0175 & 0.0235 & 0.0100 & 0.0055 & 0.0069 & 0.0084 & 0.0108 \\
&CoSeRec & 0.0099 & 0.0217 & 0.0284 & 0.0357 & 0.0154 & 0.0099 & 0.0159 & 0.0181 & 0.0199 \\
&$C^2DSR$ & 0.0155 & 0.0275 & 0.0320 & 0.0415 & 0.0229 & 0.0155 & 0.0219 & 0.0234 & 0.0257 \\

\midrule
\multirow{4}{*}{\makecell{LLM-based \\Baselines}}&UniSRec & 0.0145 & 0.0245 & 0.0375 & 0.0525 & 0.0237 & 0.0145 & 0.0201 & 0.0242 & 0.0256 \\
&BIGRec & 0.0220 & 0.0315 & 0.0370 & 0.0480 & 0.0277 & 0.0220 & 0.0268 & 0.0286 & 0.0313 \\
&GPT4Rec & 0.0165 & 0.0360 & 0.0435 & 0.0525 & 0.0262 & 0.0165 & 0.0263 & 0.0287 & 0.0311 \\
&CoLLM & 0.0170 & 0.0347 & 0.0438 & 0.0493 & 0.0286 & 0.0170 & 0.0263 & 0.0295 & 0.0309 \\
\midrule

Our Work&URLLM & \textbf{0.0270} & \textbf{0.0400} & \textbf{0.0485} & \textbf{0.0595} & \textbf{0.0355} & \textbf{0.0270} & \textbf{0.0353} & \textbf{0.0371} & \textbf{0.0399} \\
\bottomrule
\end{tabular}
\end{table*} 
We compare the performance of URLLM with state-of-the-art traditional and LLM-based CDSR methods to showcase its effectiveness. The baselines are as follows:
\subsubsection{Traditional Baselines} 

\begin{itemize}

     \item \textbf{LightGCN}~\cite{DBLP:conf/sigir/0001DWLZ020} simplifies the design of Graph Convolution Networks for collaborative filtering by focusing on the essential component of neighborhood aggregation.
     \item \textbf{SASRec}~\cite{DBLP:conf/icdm/KangM18} uses a causal attention mechanism to model sequential patterns.
     \item \textbf{CoSeRec}~\cite{DBLP:journals/corr/abs-2108-06479} introduces two new informative augmentation operators that leverage item correlations with contrastive sequence.
     \item 
    $\bm{C^2DSR}$~\cite{DBLP:conf/cikm/CaoCSLW22} captures user preferences by simultaneously leveraging intra- and inter-sequence item relationships through contrasive item-item graphs.
 \end{itemize}

\subsubsection{LLM-based Baselines}
\begin{itemize}
     \item \textbf{UniSRec}~\cite{DBLP:conf/kdd/HouMZLDW22} incorporates a lightweight item encoding architecture and employs contrastive pre-training tasks to learn transferable representations across domains.
     \item \textbf{GPT4Rec-LLaMA2}~\cite{DBLP:conf/ecom/LiZWXLM23} uses BM25 grounding method to refine answer into actual item space. We replaced the GPT2 with LLaMA2 to ensure a fair comparison.
     \item \textbf{BIGRec}~\cite{DBLP:journals/corr/abs-2308-08434} grounds recommendation space of tuned LLM into real item space by incorporating statistical information.
     \item \textbf{CoLLM}~\cite{DBLP:journals/corr/abs-2310-19488}\footnote{We transformed the item rating model into an item recommendation model by providing 5000 candidates and ranking them based on scores.} captures collaborative information using an external traditional model and maps it into LLM by adapter.
\end{itemize}
\subsection{Evaluation Metrics}
Following previous works~\cite{DBLP:conf/icdm/KangM18,DBLP:conf/cikm/SunLWPLOJ19,DBLP:conf/cikm/ZhouWZZWZWW20}, we leverage the leave-one-out method to calculate the recommendation performance. Besides, we adopt the whole item set as the candidate item set during evaluation to avoid the sampling bias of the candidate selection~\cite{krichene2020sampled}. Then, we evaluate the Top-K recommendation performance by Mean Reciprocal Rank (MRR)~\cite{DBLP:conf/trec/Voorhees99}, Normalized Discounted Cumulative Gain (NDCG)~\cite{DBLP:journals/tois/JarvelinK02} and
Hit Rate (HR)~\cite{waters1976hit}. 

\subsection{Implementation Details} 

The instruction-tuning and model inference are conducted on 4 Tesla A100 40G GPUs, which takes approximately 24 hours for training stage and 1 hour for inferencing. On our dual graph sequence-modeling model, we set $d_s=128$ and $d_a=32$ according to their scale and sparsity. $\gamma$ and $\lambda$ are all set to 0.3 for balance. Across all generative LLMs, we finetune LLaMA2-7B-chat with LoRA\cite{hu2021lora} with LoRA-rank=8 and LoRA-alpha=16 using Adam\cite{kingma2017adam} in a default learning rate of 1e-4. In the item attribute gaining part, in order to gain extra information, we use gpt-3.5-turbo-061 to gain introduction and attribute of the item.

\section{Results and Analysis}

\subsection{Overall Performance}

In this subsection, we compare the overall performance of all methods, which is presented in Table~\ref{tab:GM_datasets_benchmark} and Table~\ref{tab:AO_datasets_benchmark}. We can draw the following conclusions from the results: 1). LLM-based CDSR methods generally surpass traditional methods, particularly in the sparser Movie-Game domain. This highlights the LLMs' ability to perform extensive reasoning and few-shot learning. 2). Discriminative LLM-based method UniSRec underperforms compared to other generative LLMs, suggesting emergent reasoning capabilities in generative models. 3). In the Art-Office domain, LLMs incorporating collaborative information (BIGRec, CoLLM) outperform methods without such information (GPT4Rec), underscoring its importance in LLMs. The opposite trend exists in the sparse domain Movie-Game, indicating that inappropriate information integration has a detrimental effect on performance. 4). URLLM consistently outperforms all baselines, demonstrating its effectiveness. Its superiority over LLM-based baselines confirms the value of seamless information integration and domain-specific generation.

\begin{figure*}[t]
  \centering
  \includegraphics[width=0.95\linewidth]{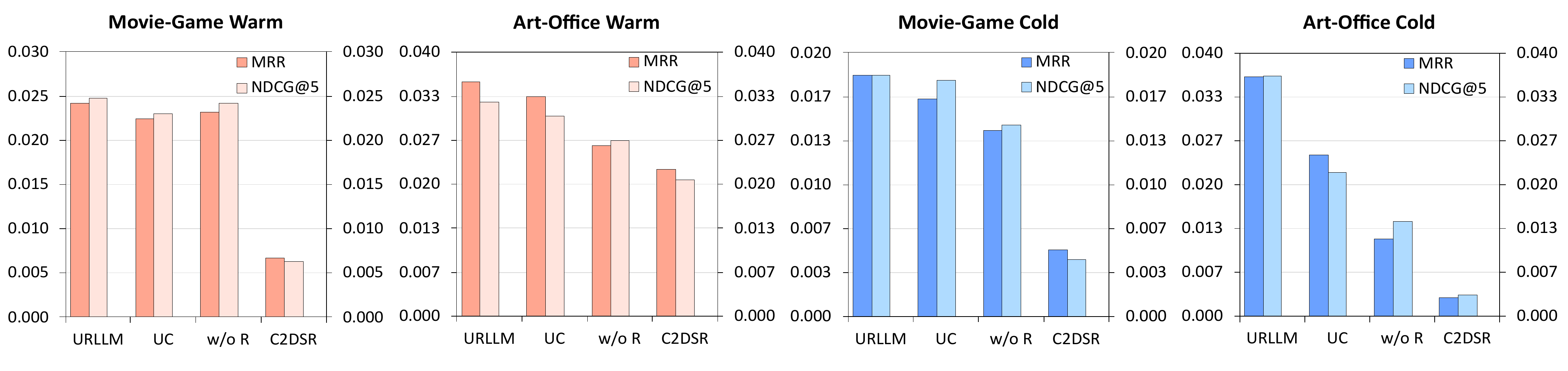}
  \caption{Performance comparison is conducted in warm (left) and cold (right) scenarios for Movie-Game and Art-Office. "UC" denotes the substitution of the retrieval model of URLLM with $\bm{C^2DSR}$, and "w/o R" designates URLLM without user retrieval.}
  \label{fig:cold_warm_start}
\end{figure*}
\begin{table*}[t]
\centering
\caption{Ablation study on Movie-Game dataset. URLLM-SASRec, URLLM-$\bm{C^2DSR}$ denotes the substitution of the retrieval model of URLLM with SASRec or $\bm{C^2DSR}$, simultaneously represent URLLM w/o item graph or URLLM w/o attribute graph.}
\begin{tabular}{l|C{1.1cm}C{1.1cm}C{1.1cm}C{1.1cm}|C{1.1cm}|C{1.1cm}C{1.1cm}C{1.1cm}C{1.1cm}}
\toprule
\textbf{Method} & \textbf{HR1} & \textbf{HR5} & \textbf{HR10} & \textbf{HR20} & \textbf{MRR} & \textbf{NG1} & \textbf{NG5} & \textbf{NG10} & \textbf{NG20} \\
\midrule
w/o LLM & 0.0069 & 0.0091 & 0.0108 & 0.0127 & 0.0082 & 0.0069 & 0.0080 & 0.0085 & 0.0090 \\
w/o retrieve & 0.0088 & 0.0322 & 0.0410 & 0.0515 & 0.0197 & 0.0088 & 0.0208 & 0.0234 & 0.0266 \\
w/o graph alignment & 0.0102 & 0.0331 & 0.0414 & 0.0522 & 0.0210 & 0.0102 & 0.0208 & 0.0243 & 0.0268 \\
w/o answer refine & 0.0105 & 0.0302 & 0.0410 & 0.0510 & 0.0203 & 0.0105 & 0.0206 & 0.0241 & 0.0267 \\
w/o domain-specific retrieval&0.0074&	0.0197&	0.0269&	0.0366&	0.0143&	0.0074	&0.0131	&0.0155&	0.0180\\

URLLM-SASRec & 0.0099&	0.0308&	0.0409&	0.0505&	0.0201&	0.0099&	0.0208&	0.0244&	0.0266\\
URLLM-$C^2DSR$ & 0.0089 & 0.0313 & 0.0402 & 0.0489 & 0.0190 & 0.0089 & 0.0201 & 0.0229 & 0.0251 \\
URLLM & \textbf{0.0105} & \textbf{0.0333} & \textbf{0.0416} & \textbf{0.0522} & \textbf{0.0211} & \textbf{0.0105} & \textbf{0.0221} & \textbf{0.0248} & \textbf{0.0298} \\
\bottomrule
\end{tabular}
  \label{tab:ablasion_GM}
\end{table*}

\begin{table*}[t]
\centering
\caption{Ablation study on Art-Office dataset. URLLM-SASRec, URLLM-$\bm{C^2DSR}$ denotes the substitution of the retrieval model of URLLM with SASRec or $\bm{C^2DSR}$, simultaneously represent URLLM w/o item graph or URLLM w/o attribute graph.}
\begin{tabular}{l|C{1.1cm}C{1.1cm}C{1.1cm}C{1.1cm}|C{1.1cm}|C{1.1cm}C{1.1cm}C{1.1cm}C{1.1cm}}
\toprule
\textbf{Method} & \textbf{HR1} & \textbf{HR5} & \textbf{HR10} & \textbf{HR20} & \textbf{MRR} & \textbf{NG1} & \textbf{NG5} & \textbf{NG10} & \textbf{NG20} \\
\midrule
w/o LLM & 0.0265 & 0.0345 & 0.0430 & 0.0505 & 0.0322 & 0.0265 & 0.0308 & 0.0336 & 0.0355 \\
w/o retriever & 0.0165 & 0.0360 & 0.0435 & 0.0525 & 0.0262 & 0.0165 & 0.0263 & 0.0287 & 0.0311 \\
w/o graph alignment & 0.0235 & 0.0415 & 0.0475 & 0.0530 & 0.0322 & 0.0235 & 0.0324 & 0.0344 & 0.0357 \\
w/o answer refine & 0.0230 & 0.0415 & 0.0480 & 0.0560 & 0.0323 & 0.0230 & 0.0329 & 0.0349 & 0.0370 \\
w/o domain-specific retrieval&0.0235&	0.0315&	0.0365&	0.0455&	0.0289&	0.0235&	0.0274	&0.0292&	0.0307\\
URLLM-SASRec & 0.0175 & 0.0350 & 0.0420 & 0.0490 & 0.0263 & 0.0175 & 0.0267 & 0.0290 & 0.0307 \\
URLLM-$C^2DSR$ & 0.0215&	0.0375&0.0455	&0.0550&	0.0302&	0.0215	&0.0309	&0.0328&	0.0349
 \\
URLLM & \textbf{0.0270} & \textbf{0.0430} & \textbf{0.0485} & \textbf{0.0595} & \textbf{0.0355} & \textbf{0.0270} & \textbf{0.0353} & \textbf{0.0371} & \textbf{0.0399} \\
\bottomrule
\end{tabular}
  \label{tab:ablasion_AO}
\end{table*}

\subsection{Analysis on warm/cold start scenario}

URLLM seeks to retrieve similar users into LLM, aiming to integrate additional knowledge tackling cold-start scenarios of CDSR. To evaluate the achievement of this goal, we conduct a detailed examination of the methods’ performance in both warm and cold-start scenarios. In particular, users are categorized into cold scenarios when the length of their domain-relevant interaction sequences is less than 3. Other users are categorized into warm scenarios. To showcase the efficacy of our model in cold-start situations, comparison among four models are compared in Figure~\ref{fig:cold_warm_start}. We can draw three pieces of information:
1). In comparison to the model devoid of the retrieval module (w/o R), URLLM exhibits an enhancement in cold-start scenarios across both datasets. This underscores the efficacy of the user-retrieval module in augmenting the model's performance. Notably, in cold-start scenarios on the Art-Office dataset, our model surpasses its warm-start performance (0.0364 versus 0.0355 in MRR). This suggests that URLLM has the potential to convert cold-start scenarios into warm-start ones.

2). 
We observe that, within the context of the movie-game dataset's warm-start setting, the performance of user retrieval based on $C^2DSR$ is inferior to not performing any retrieval. This is attributed to $C^2DSR$'s inability to model structural-semantic information, which consequently impairs its user retrieval capability. We hypothesize that this deficiency stems from the absence of semantic information modeling, introducing noise into LLMs, particularly under warm-start conditions.

3). URLLM's efficacy in cold-start scenarios is demonstrably superior compared to the traditional $C^2DSR$ model. On the Movie-Game dataset, URLLM achieves a remarkable 3.6-fold improvement, while on the Art-Office dataset, the performance gap widens to a factor of 13. These substantial gains highlight URLLM's effectiveness in augmenting the model's capabilities for dealing with sparse data and limited user interaction.

\subsection{Ablation study}

To demonstrate the effectiveness of each component, we conduct an ablation study to compare URLLM with seven variants: (1) w/o LLM: removing LLM to use only dual graph sequence modeling model to recommend; (2) w/o retrieve: removing retriever to adopt LLaMA2 itself; (3) w/o graph alignment: removing the alignment loss by setting $\gamma=0$; (4) w/o answer refine: generating without domain-specific refine; (5) w/o domain-specific retrieval: retrieving users without domain differentiation strategy; (6) URLLM-SASRec: replacing dual graph sequence modeling model with SASRec; (7) URLLM-$C^2DSR$: replacing dual graph sequence modeling model with $C^2DSR$. The results are presented in Table~\ref{tab:ablasion_GM} and Table~\ref{tab:ablasion_AO}.

The results demonstrate that each component within our framework plays a positive role in enhancing overall performance. Notably, for the Art-Office data, removing the retrieval and answer refinement module significantly decreased recommendation effectiveness. This highlights the critical role of collaborative information, employed in both retrieval and refining. Conversely, in the sparser Movie-Game dataset, the LLM and attribute graph exhibited the most significant contributions, suggesting that semantic information is paramount for dealing with cold-start scenarios.
 \begin{table*}[t]
\centering
\caption{ Movie-Game CDSR performance replacing on user retrieval with candidates. The subscript model in the Method denotes the application of the model for either user retrieval or the generation of top-k candidates.}
\label{tab:cd_replace_GM}

\begin{tabular}{l|c|ccccccccc}
\toprule
\textbf{Method} & \textbf{UHR($10^{-3}$)} & \textbf{HR@1} & \textbf{HR@5} & \textbf{HR@10} & \textbf{HR@20} & \textbf{MRR} & \textbf{NG@1} & \textbf{NG@5} & \textbf{NG@10} & \textbf{NG@20} \\
\hline
$Candidate_{C^2DSR}$ & 0.4700 & 0.0083 & 0.0277 & 0.0369 & 0.0450 & 0.0180 & 0.0083 & 0.0183 & 0.0213 & 0.0232 \\
$Retrieval_{C^2DSR}$ & 0.3694 & 0.0089 & 0.0313 & 0.0402 & 0.0489 & 0.0190 & 0.0089 & 0.0201 & 0.0229 & 0.0251 \\
$Candidate_{URLLM}$ & \textbf{0.6350} & 0.0089 & 0.0275 & 0.0378 & 0.4580 & 0.0184 & 0.0089 & 0.0185 & 0.0219 & 0.0240 \\
$Retrieval_{URLLM}$ & 0.5651 & \textbf{0.0105} & \textbf{0.0333} & \textbf{0.0416} & \textbf{0.0522} & \textbf{0.0211} & \textbf{0.0105} & \textbf{0.0221} & \textbf{0.0248} & \textbf{0.0298} \\
\bottomrule
\end{tabular}
\end{table*}

\begin{table*}[t]

\centering
\caption{  Art-Office CDSR performance replacing on user retrieval with candidates. The subscript model in the Method denotes the application of the model for either user retrieval or the generation of top-k candidates.
}
\label{tab:cd_replace_AO}
\begin{tabular}{l|c|ccccccccc}
\toprule
\textbf{Method} & \textbf{UHR($10^{-3}$)} & \textbf{HR@1} & \textbf{HR@5} & \textbf{HR@10} & \textbf{HR@20} & \textbf{MRR} & \textbf{NG@1} & \textbf{NG@5} & \textbf{NG@10} & \textbf{NG@20} \\
\hline
$Candidate_{C^2DSR}$ & 1.5992 & 0.0185 & 0.0325 & 0.0370 & 0.0440 & 0.0257 & 0.0185 & 0.0260 & 0.0274 & 0.0292 \\
$Retrieval_{C^2DSR}$ & 0.8783 & 0.0215 & 0.0375 & 0.0455 & 0.0550 & 0.0302 & 0.0215 & 0.0309 & 0.0328 & 0.0349 \\
$Candidate_{URLLM}$ & \textbf{2.1511} & 0.0200 & 0.0340 & 0.0495 & 0.0455 & 0.0274 & 0.0200 & 0.0277 & 0.0295 & 0.0310 \\
$Retrieval_{URLLM}$ & 1.6876 & \textbf{0.0270} & \textbf{0.0430} & \textbf{0.0485} & \textbf{0.0595} & \textbf{0.0355} & \textbf{0.0270} & \textbf{0.0353} & \textbf{0.0371} & \textbf{0.0399} \\
\bottomrule
\end{tabular}
\end{table*}

Furthermore, when we dissociate the domain-specific retrieval method or substitute the Dual Graph Sequence Modeling Model with SASRec and $C^2DSR$—which fail to generate user representation in specific domains—the model's performance deteriorates due to the loss of crucial collaborative information and the integration of noise. The answer refinement model also affects the results of the model as shown in Table~\ref{tab:ablasion_GM} and Table~\ref{tab:ablasion_AO}, which shows the importance of domain-specific generation.

\subsection{Analysis on integration of LLM}

\subsubsection{Analysis on quality of user integration}
In general, superior-performing models tend to produce more precise user representations during training, which in turn leads to improved user retrieval results. 
Nevertheless, there appears to be an inconsistency between the model performance depicted in 
Table~\ref{tab:GM_datasets_benchmark} and the retrieval outcomes presented in Table~\ref{tab:ablasion_GM}.
While $C^2DSR$ outperforms SASRec in terms of retrieval performance, its effectiveness diminishes when used as a replacement for the search model than SASRec.
This observation prompts us to explore the impact of the retrieval result quality on recommendation performance.

\begin{figure}[h]
  \centering
  \includegraphics[width=0.95\linewidth]{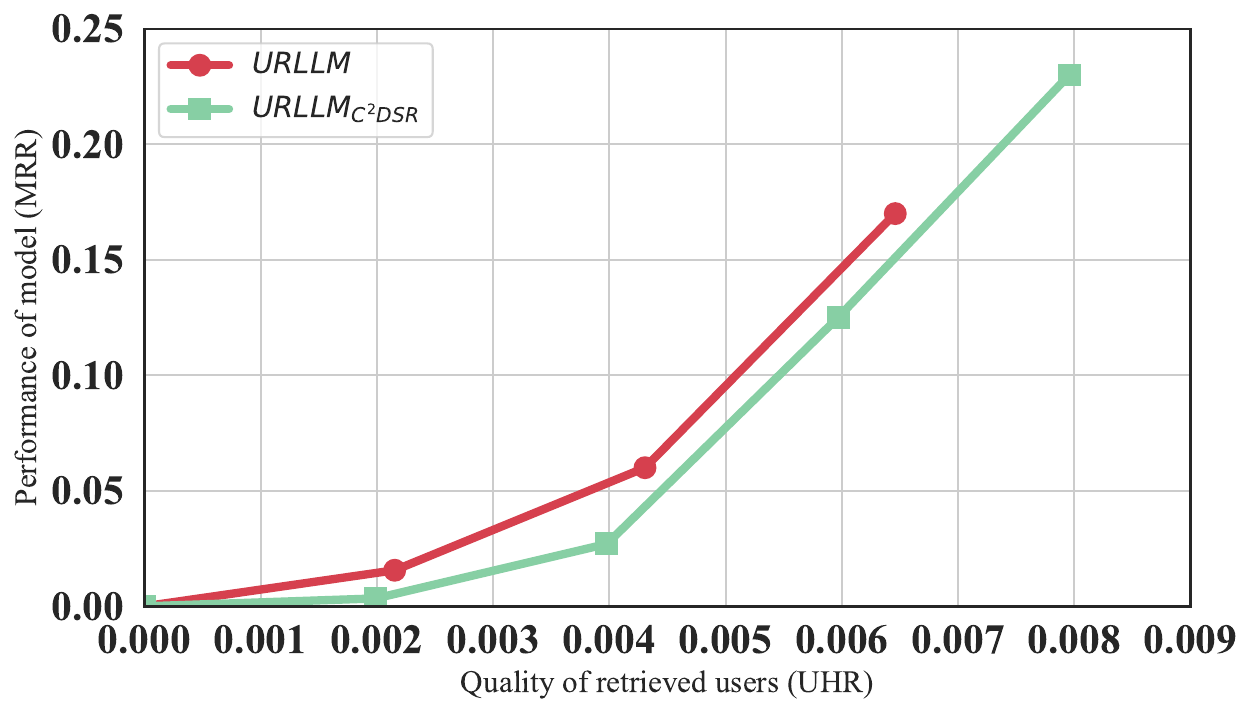}
  \caption{The positive correlation between quality of retrieved user and performance of model.}
  \label{fig:HRT-MRR}
\end{figure}

To quantitatively assess performance, we employ User Hit Rate (UHR) dividing the hit rate of the user interaction by the interaction length for retrieval quality and Mean Reciprocal Rank (MRR) for recommendation effectiveness. We further select a diverse subset of Art-Office examples, encompassing both higher-quality and lower-quality retrievals. Figure~\ref{fig:HRT-MRR} illustrates the observed relationship between retrieval result quality and recommendation performance. Notably, this experiment is not conducted on the Movie-Game dataset due to its sparsity, which leads to unstable retrieval results and compromises the reliability of sampled outcomes.

Figure~\ref{fig:HRT-MRR} reveals that URLLM outperforms $C^2DSR$, suggesting its superior ability to model collaborative information. However, the close alignment of their curves indicates that model choice becomes less impactful. Notably, the positive correlation between MRR and UHR underscores the influence of retrieval quality on performance.

Therefore, we can now address the initial question posed in this subsection: even though a stronger model leads to $HR_{SASRec}=0.0067 < HR_{C^2DSR}=0.0097$, the length of retrieved user $l_{SASRec}=14.41 < l_{C^2DSR}=26.26$ resulting in $UHR_1/UHR_2=1.24>1$, this causes a reduction of the $C^2DSR$ retrieval model.

\subsubsection{Analysis on other linguistic integration of user integration}

We replaced user retrieval with candidate provision to explore more semantic integration approaches, ensuring that their lengths are equal. The model's performance using candidates can be observed in Table~\ref{tab:cd_replace_GM} and Table~\ref{tab:cd_replace_AO}. It is noteworthy that, even though the UHR value for candidates is higher than our user retrieval under the same length, their overall final performance is inferior to our model. Additionally, in the experiments, we discovered some interesting phenomena – In a limited subset of instances even when the user retrieval UHR is 0, our model's performance still improves. This demonstrates the importance of collaborative information in user retrieval and validates the correctness of our approach.

\section{Conclusion}
In conclusion, this paper introduced URLLM, a novel user retrieval CDSR framework to seamlessly incorporate diverse information into LLM. Initially, we developed a dual-graph sequence-modeling framework to capture collaborative and structural-semantic information derived from user interactions. Subsequently, we devised a novel user retrieval model for LLMs, aimed at infusing this knowledge into LLMs by exploiting their robust language reasoning and ensemble capabilities. Furthermore, the domain-specific retrieval strategy and answer refinement module were proposed for domain-specific information integration and generation.

Compared to traditional approaches and other LLM-based methods, URLLM exhibited improved performance on the CDSA task. This research not only explored the relationship between provided user retrieval results and model performance but also propelled LLM-based CDSR toward the desired format. In further work, we will attempt to retrench the retrieved user length and evaluations on larger-scale models.

\bibliographystyle{ACM-Reference-Format}
\bibliography{citation}


\begin{thebibliography}{66}


\ifx \showCODEN    \undefined \def \showCODEN     #1{\unskip}     \fi
\ifx \showDOI      \undefined \def \showDOI       #1{#1}\fi
\ifx \showISBNx    \undefined \def \showISBNx     #1{\unskip}     \fi
\ifx \showISBNxiii \undefined \def \showISBNxiii  #1{\unskip}     \fi
\ifx \showISSN     \undefined \def \showISSN      #1{\unskip}     \fi
\ifx \showLCCN     \undefined \def \showLCCN      #1{\unskip}     \fi
\ifx \shownote     \undefined \def \shownote      #1{#1}          \fi
\ifx \showarticletitle \undefined \def \showarticletitle #1{#1}   \fi
\ifx \showURL      \undefined \def \showURL       {\relax}        \fi
\providecommand\bibfield[2]{#2}
\providecommand\bibinfo[2]{#2}
\providecommand\natexlab[1]{#1}
\providecommand\showeprint[2][]{arXiv:#2}

\bibitem[Alharbi and Caragea(2022)]%
        {10.1007/978-981-16-5348-3_48}
\bibfield{author}{\bibinfo{person}{Nawaf Alharbi} {and} \bibinfo{person}{Doina Caragea}.} \bibinfo{year}{2022}\natexlab{}.
\newblock \showarticletitle{Cross-domain Self-attentive Sequential Recommendations}. In \bibinfo{booktitle}{\emph{Proceedings of International Conference on Data Science and Applications}}, \bibfield{editor}{\bibinfo{person}{Mukesh Saraswat}, \bibinfo{person}{Sarbani Roy}, \bibinfo{person}{Chandreyee Chowdhury}, {and} \bibinfo{person}{Amir~H. Gandomi}} (Eds.). \bibinfo{publisher}{Springer Singapore}, \bibinfo{address}{Singapore}, \bibinfo{pages}{601--614}.
\newblock
\showISBNx{978-981-16-5348-3}


\bibitem[Bao et~al\mbox{.}(2023a)]%
        {bao2023bistep}
\bibfield{author}{\bibinfo{person}{Keqin Bao}, \bibinfo{person}{Jizhi Zhang}, \bibinfo{person}{Wenjie Wang}, \bibinfo{person}{Yang Zhang}, \bibinfo{person}{Zhengyi Yang}, \bibinfo{person}{Yancheng Luo}, \bibinfo{person}{Chong Chen}, \bibinfo{person}{Fuli Feng}, {and} \bibinfo{person}{Qi Tian}.} \bibinfo{year}{2023}\natexlab{a}.
\newblock \bibinfo{title}{A Bi-Step Grounding Paradigm for Large Language Models in Recommendation Systems}.
\newblock
\newblock
\showeprint[arxiv]{2308.08434}~[cs.IR]


\bibitem[Bao et~al\mbox{.}(2023b)]%
        {DBLP:journals/corr/abs-2308-08434}
\bibfield{author}{\bibinfo{person}{Keqin Bao}, \bibinfo{person}{Jizhi Zhang}, \bibinfo{person}{Wenjie Wang}, \bibinfo{person}{Yang Zhang}, \bibinfo{person}{Zhengyi Yang}, \bibinfo{person}{Yancheng Luo}, \bibinfo{person}{Fuli Feng}, \bibinfo{person}{Xiangnan He}, {and} \bibinfo{person}{Qi Tian}.} \bibinfo{year}{2023}\natexlab{b}.
\newblock \showarticletitle{A Bi-Step Grounding Paradigm for Large Language Models in Recommendation Systems}.
\newblock \bibinfo{journal}{\emph{CoRR}}  \bibinfo{volume}{abs/2308.08434} (\bibinfo{year}{2023}).
\newblock
\urldef\tempurl%
\url{https://doi.org/10.48550/ARXIV.2308.08434}
\showDOI{\tempurl}
\showeprint[arXiv]{2308.08434}


\bibitem[Bao et~al\mbox{.}(2023c)]%
        {10.1145/3604915.3608857}
\bibfield{author}{\bibinfo{person}{Keqin Bao}, \bibinfo{person}{Jizhi Zhang}, \bibinfo{person}{Yang Zhang}, \bibinfo{person}{Wenjie Wang}, \bibinfo{person}{Fuli Feng}, {and} \bibinfo{person}{Xiangnan He}.} \bibinfo{year}{2023}\natexlab{c}.
\newblock \showarticletitle{TALLRec: An Effective and Efficient Tuning Framework to Align Large Language Model with Recommendation}. In \bibinfo{booktitle}{\emph{Proceedings of the 17th ACM Conference on Recommender Systems}} (Singapore, Singapore) \emph{(\bibinfo{series}{RecSys '23})}. \bibinfo{publisher}{Association for Computing Machinery}, \bibinfo{address}{New York, NY, USA}, \bibinfo{pages}{1007–1014}.
\newblock
\showISBNx{9798400702419}
\urldef\tempurl%
\url{https://doi.org/10.1145/3604915.3608857}
\showDOI{\tempurl}


\bibitem[Boiko et~al\mbox{.}(2023)]%
        {DBLP:journals/corr/abs-2304-05332}
\bibfield{author}{\bibinfo{person}{Daniil~A. Boiko}, \bibinfo{person}{Robert MacKnight}, {and} \bibinfo{person}{Gabe Gomes}.} \bibinfo{year}{2023}\natexlab{}.
\newblock \showarticletitle{Emergent autonomous scientific research capabilities of large language models}.
\newblock \bibinfo{journal}{\emph{CoRR}}  \bibinfo{volume}{abs/2304.05332} (\bibinfo{year}{2023}).
\newblock
\urldef\tempurl%
\url{https://doi.org/10.48550/ARXIV.2304.05332}
\showDOI{\tempurl}
\showeprint[arXiv]{2304.05332}


\bibitem[Cao et~al\mbox{.}(2022a)]%
        {10.1145/3511808.3557262}
\bibfield{author}{\bibinfo{person}{Jiangxia Cao}, \bibinfo{person}{Xin Cong}, \bibinfo{person}{Jiawei Sheng}, \bibinfo{person}{Tingwen Liu}, {and} \bibinfo{person}{Bin Wang}.} \bibinfo{year}{2022}\natexlab{a}.
\newblock \showarticletitle{Contrastive Cross-Domain Sequential Recommendation}. In \bibinfo{booktitle}{\emph{Proceedings of the 31st ACM International Conference on Information \& Knowledge Management}} (Atlanta, GA, USA) \emph{(\bibinfo{series}{CIKM '22})}. \bibinfo{publisher}{Association for Computing Machinery}, \bibinfo{address}{New York, NY, USA}, \bibinfo{pages}{138–147}.
\newblock
\showISBNx{9781450392365}
\urldef\tempurl%
\url{https://doi.org/10.1145/3511808.3557262}
\showDOI{\tempurl}


\bibitem[Cao et~al\mbox{.}(2022b)]%
        {DBLP:conf/cikm/CaoCSLW22}
\bibfield{author}{\bibinfo{person}{Jiangxia Cao}, \bibinfo{person}{Xin Cong}, \bibinfo{person}{Jiawei Sheng}, \bibinfo{person}{Tingwen Liu}, {and} \bibinfo{person}{Bin Wang}.} \bibinfo{year}{2022}\natexlab{b}.
\newblock \showarticletitle{Contrastive Cross-Domain Sequential Recommendation}. In \bibinfo{booktitle}{\emph{Proceedings of the 31st {ACM} International Conference on Information {\&} Knowledge Management, Atlanta, GA, USA, October 17-21, 2022}}, \bibfield{editor}{\bibinfo{person}{Mohammad~Al Hasan} {and} \bibinfo{person}{Li~Xiong}} (Eds.). \bibinfo{publisher}{{ACM}}, \bibinfo{pages}{138--147}.
\newblock
\urldef\tempurl%
\url{https://doi.org/10.1145/3511808.3557262}
\showDOI{\tempurl}


\bibitem[Chen(2023)]%
        {DBLP:journals/corr/abs-2311-12338}
\bibfield{author}{\bibinfo{person}{Junyi Chen}.} \bibinfo{year}{2023}\natexlab{}.
\newblock \showarticletitle{A Survey on Large Language Models for Personalized and Explainable Recommendations}.
\newblock \bibinfo{journal}{\emph{CoRR}}  \bibinfo{volume}{abs/2311.12338} (\bibinfo{year}{2023}).
\newblock
\urldef\tempurl%
\url{https://doi.org/10.48550/ARXIV.2311.12338}
\showDOI{\tempurl}
\showeprint[arXiv]{2311.12338}


\bibitem[Fan et~al\mbox{.}(2023)]%
        {DBLP:journals/corr/abs-2307-02046}
\bibfield{author}{\bibinfo{person}{Wenqi Fan}, \bibinfo{person}{Zihuai Zhao}, \bibinfo{person}{Jiatong Li}, \bibinfo{person}{Yunqing Liu}, \bibinfo{person}{Xiaowei Mei}, \bibinfo{person}{Yiqi Wang}, \bibinfo{person}{Jiliang Tang}, {and} \bibinfo{person}{Qing Li}.} \bibinfo{year}{2023}\natexlab{}.
\newblock \showarticletitle{Recommender Systems in the Era of Large Language Models (LLMs)}.
\newblock \bibinfo{journal}{\emph{CoRR}}  \bibinfo{volume}{abs/2307.02046} (\bibinfo{year}{2023}).
\newblock
\urldef\tempurl%
\url{https://doi.org/10.48550/ARXIV.2307.02046}
\showDOI{\tempurl}
\showeprint[arXiv]{2307.02046}


\bibitem[Gao et~al\mbox{.}(2023)]%
        {DBLP:journals/corr/abs-2303-14524}
\bibfield{author}{\bibinfo{person}{Yunfan Gao}, \bibinfo{person}{Tao Sheng}, \bibinfo{person}{Youlin Xiang}, \bibinfo{person}{Yun Xiong}, \bibinfo{person}{Haofen Wang}, {and} \bibinfo{person}{Jiawei Zhang}.} \bibinfo{year}{2023}\natexlab{}.
\newblock \showarticletitle{Chat-REC: Towards Interactive and Explainable LLMs-Augmented Recommender System}.
\newblock \bibinfo{journal}{\emph{CoRR}}  \bibinfo{volume}{abs/2303.14524} (\bibinfo{year}{2023}).
\newblock
\urldef\tempurl%
\url{https://doi.org/10.48550/ARXIV.2303.14524}
\showDOI{\tempurl}
\showeprint[arXiv]{2303.14524}


\bibitem[Geng et~al\mbox{.}(2022)]%
        {10.1145/3523227.3546767}
\bibfield{author}{\bibinfo{person}{Shijie Geng}, \bibinfo{person}{Shuchang Liu}, \bibinfo{person}{Zuohui Fu}, \bibinfo{person}{Yingqiang Ge}, {and} \bibinfo{person}{Yongfeng Zhang}.} \bibinfo{year}{2022}\natexlab{}.
\newblock \showarticletitle{Recommendation as Language Processing (RLP): A Unified Pretrain, Personalized Prompt \& Predict Paradigm (P5)}. In \bibinfo{booktitle}{\emph{Proceedings of the 16th ACM Conference on Recommender Systems}} (Seattle, WA, USA) \emph{(\bibinfo{series}{RecSys '22})}. \bibinfo{publisher}{Association for Computing Machinery}, \bibinfo{address}{New York, NY, USA}, \bibinfo{pages}{299–315}.
\newblock
\showISBNx{9781450392785}
\urldef\tempurl%
\url{https://doi.org/10.1145/3523227.3546767}
\showDOI{\tempurl}


\bibitem[Gong et~al\mbox{.}(2023)]%
        {10.1145/3583780.3614657}
\bibfield{author}{\bibinfo{person}{Yuqi Gong}, \bibinfo{person}{Xichen Ding}, \bibinfo{person}{Yehui Su}, \bibinfo{person}{Kaiming Shen}, \bibinfo{person}{Zhongyi Liu}, {and} \bibinfo{person}{Guannan Zhang}.} \bibinfo{year}{2023}\natexlab{}.
\newblock \showarticletitle{An Unified Search and Recommendation Foundation Model for Cold-Start Scenario}. In \bibinfo{booktitle}{\emph{Proceedings of the 32nd ACM International Conference on Information and Knowledge Management}} (<conf-loc>, <city>Birmingham</city>, <country>United Kingdom</country>, </conf-loc>) \emph{(\bibinfo{series}{CIKM '23})}. \bibinfo{publisher}{Association for Computing Machinery}, \bibinfo{address}{New York, NY, USA}, \bibinfo{pages}{4595–4601}.
\newblock
\showISBNx{9798400701245}
\urldef\tempurl%
\url{https://doi.org/10.1145/3583780.3614657}
\showDOI{\tempurl}


\bibitem[Guo et~al\mbox{.}(2021)]%
        {ijcai2021p342}
\bibfield{author}{\bibinfo{person}{Lei Guo}, \bibinfo{person}{Li Tang}, \bibinfo{person}{Tong Chen}, \bibinfo{person}{Lei Zhu}, \bibinfo{person}{Quoc Viet~Hung Nguyen}, {and} \bibinfo{person}{Hongzhi Yin}.} \bibinfo{year}{2021}\natexlab{}.
\newblock \showarticletitle{DA-GCN: A Domain-aware Attentive Graph Convolution Network for Shared-account Cross-domain Sequential Recommendation}. In \bibinfo{booktitle}{\emph{Proceedings of the Thirtieth International Joint Conference on Artificial Intelligence, {IJCAI-21}}}, \bibfield{editor}{\bibinfo{person}{Zhi-Hua Zhou}} (Ed.). \bibinfo{publisher}{International Joint Conferences on Artificial Intelligence Organization}, \bibinfo{pages}{2483--2489}.
\newblock
\urldef\tempurl%
\url{https://doi.org/10.24963/ijcai.2021/342}
\showDOI{\tempurl}
\newblock
\shownote{Main Track}.


\bibitem[Han et~al\mbox{.}(2024)]%
        {han2024end4rec}
\bibfield{author}{\bibinfo{person}{Yongqiang Han}, \bibinfo{person}{Hao Wang}, \bibinfo{person}{Kefan Wang}, \bibinfo{person}{Likang Wu}, \bibinfo{person}{Zhi Li}, \bibinfo{person}{Wei Guo}, \bibinfo{person}{Yong Liu}, \bibinfo{person}{Defu Lian}, {and} \bibinfo{person}{Enhong Chen}.} \bibinfo{year}{2024}\natexlab{}.
\newblock \showarticletitle{END4Rec: Efficient Noise-Decoupling for Multi-Behavior Sequential Recommendation}.
\newblock \bibinfo{journal}{\emph{arXiv preprint arXiv:2403.17603}} (\bibinfo{year}{2024}).
\newblock


\bibitem[Han et~al\mbox{.}(2023)]%
        {han2023guesr}
\bibfield{author}{\bibinfo{person}{Yongqiang Han}, \bibinfo{person}{Likang Wu}, \bibinfo{person}{Hao Wang}, \bibinfo{person}{Guifeng Wang}, \bibinfo{person}{Mengdi Zhang}, \bibinfo{person}{Zhi Li}, \bibinfo{person}{Defu Lian}, {and} \bibinfo{person}{Enhong Chen}.} \bibinfo{year}{2023}\natexlab{}.
\newblock \showarticletitle{Guesr: A global unsupervised data-enhancement with bucket-cluster sampling for sequential recommendation}. In \bibinfo{booktitle}{\emph{International Conference on Database Systems for Advanced Applications}}. Springer, \bibinfo{pages}{286--296}.
\newblock


\bibitem[He et~al\mbox{.}(2020)]%
        {DBLP:conf/sigir/0001DWLZ020}
\bibfield{author}{\bibinfo{person}{Xiangnan He}, \bibinfo{person}{Kuan Deng}, \bibinfo{person}{Xiang Wang}, \bibinfo{person}{Yan Li}, \bibinfo{person}{Yong{-}Dong Zhang}, {and} \bibinfo{person}{Meng Wang}.} \bibinfo{year}{2020}\natexlab{}.
\newblock \showarticletitle{LightGCN: Simplifying and Powering Graph Convolution Network for Recommendation}. In \bibinfo{booktitle}{\emph{Proceedings of the 43rd International {ACM} {SIGIR} conference on research and development in Information Retrieval, {SIGIR} 2020, Virtual Event, China, July 25-30, 2020}}, \bibfield{editor}{\bibinfo{person}{Jimmy~X. Huang}, \bibinfo{person}{Yi~Chang}, \bibinfo{person}{Xueqi Cheng}, \bibinfo{person}{Jaap Kamps}, \bibinfo{person}{Vanessa Murdock}, \bibinfo{person}{Ji{-}Rong Wen}, {and} \bibinfo{person}{Yiqun Liu}} (Eds.). \bibinfo{publisher}{{ACM}}, \bibinfo{pages}{639--648}.
\newblock
\urldef\tempurl%
\url{https://doi.org/10.1145/3397271.3401063}
\showDOI{\tempurl}


\bibitem[He et~al\mbox{.}(2016)]%
        {10.1145/2911451.2911489}
\bibfield{author}{\bibinfo{person}{Xiangnan He}, \bibinfo{person}{Hanwang Zhang}, \bibinfo{person}{Min-Yen Kan}, {and} \bibinfo{person}{Tat-Seng Chua}.} \bibinfo{year}{2016}\natexlab{}.
\newblock \showarticletitle{Fast Matrix Factorization for Online Recommendation with Implicit Feedback}. In \bibinfo{booktitle}{\emph{Proceedings of the 39th International ACM SIGIR Conference on Research and Development in Information Retrieval}} (Pisa, Italy) \emph{(\bibinfo{series}{SIGIR '16})}. \bibinfo{publisher}{Association for Computing Machinery}, \bibinfo{address}{New York, NY, USA}, \bibinfo{pages}{549–558}.
\newblock
\showISBNx{9781450340694}
\urldef\tempurl%
\url{https://doi.org/10.1145/2911451.2911489}
\showDOI{\tempurl}


\bibitem[Hidasi and Karatzoglou(2018)]%
        {10.1145/3269206.3271761}
\bibfield{author}{\bibinfo{person}{Bal\'{a}zs Hidasi} {and} \bibinfo{person}{Alexandros Karatzoglou}.} \bibinfo{year}{2018}\natexlab{}.
\newblock \showarticletitle{Recurrent Neural Networks with Top-k Gains for Session-based Recommendations}. In \bibinfo{booktitle}{\emph{Proceedings of the 27th ACM International Conference on Information and Knowledge Management}} (Torino, Italy) \emph{(\bibinfo{series}{CIKM '18})}. \bibinfo{publisher}{Association for Computing Machinery}, \bibinfo{address}{New York, NY, USA}, \bibinfo{pages}{843–852}.
\newblock
\showISBNx{9781450360142}
\urldef\tempurl%
\url{https://doi.org/10.1145/3269206.3271761}
\showDOI{\tempurl}


\bibitem[Hidasi et~al\mbox{.}(2016)]%
        {DBLP:journals/corr/HidasiKBT15}
\bibfield{author}{\bibinfo{person}{Bal{\'{a}}zs Hidasi}, \bibinfo{person}{Alexandros Karatzoglou}, \bibinfo{person}{Linas Baltrunas}, {and} \bibinfo{person}{Domonkos Tikk}.} \bibinfo{year}{2016}\natexlab{}.
\newblock \showarticletitle{Session-based Recommendations with Recurrent Neural Networks}. In \bibinfo{booktitle}{\emph{4th International Conference on Learning Representations, {ICLR} 2016, San Juan, Puerto Rico, May 2-4, 2016, Conference Track Proceedings}}, \bibfield{editor}{\bibinfo{person}{Yoshua Bengio} {and} \bibinfo{person}{Yann LeCun}} (Eds.).
\newblock
\urldef\tempurl%
\url{http://arxiv.org/abs/1511.06939}
\showURL{%
\tempurl}


\bibitem[Hou et~al\mbox{.}(2022a)]%
        {DBLP:conf/kdd/HouMZLDW22}
\bibfield{author}{\bibinfo{person}{Yupeng Hou}, \bibinfo{person}{Shanlei Mu}, \bibinfo{person}{Wayne~Xin Zhao}, \bibinfo{person}{Yaliang Li}, \bibinfo{person}{Bolin Ding}, {and} \bibinfo{person}{Ji{-}Rong Wen}.} \bibinfo{year}{2022}\natexlab{a}.
\newblock \showarticletitle{Towards Universal Sequence Representation Learning for Recommender Systems}. In \bibinfo{booktitle}{\emph{{KDD} '22: The 28th {ACM} {SIGKDD} Conference on Knowledge Discovery and Data Mining, Washington, DC, USA, August 14 - 18, 2022}}, \bibfield{editor}{\bibinfo{person}{Aidong Zhang} {and} \bibinfo{person}{Huzefa Rangwala}} (Eds.). \bibinfo{publisher}{{ACM}}, \bibinfo{pages}{585--593}.
\newblock
\urldef\tempurl%
\url{https://doi.org/10.1145/3534678.3539381}
\showDOI{\tempurl}


\bibitem[Hou et~al\mbox{.}(2022b)]%
        {10.1145/3534678.3539381}
\bibfield{author}{\bibinfo{person}{Yupeng Hou}, \bibinfo{person}{Shanlei Mu}, \bibinfo{person}{Wayne~Xin Zhao}, \bibinfo{person}{Yaliang Li}, \bibinfo{person}{Bolin Ding}, {and} \bibinfo{person}{Ji-Rong Wen}.} \bibinfo{year}{2022}\natexlab{b}.
\newblock \showarticletitle{Towards Universal Sequence Representation Learning for Recommender Systems}. In \bibinfo{booktitle}{\emph{Proceedings of the 28th ACM SIGKDD Conference on Knowledge Discovery and Data Mining}} (Washington DC, USA) \emph{(\bibinfo{series}{KDD '22})}. \bibinfo{publisher}{Association for Computing Machinery}, \bibinfo{address}{New York, NY, USA}, \bibinfo{pages}{585–593}.
\newblock
\showISBNx{9781450393850}
\urldef\tempurl%
\url{https://doi.org/10.1145/3534678.3539381}
\showDOI{\tempurl}


\bibitem[Hu et~al\mbox{.}(2021)]%
        {hu2021lora}
\bibfield{author}{\bibinfo{person}{Edward~J Hu}, \bibinfo{person}{Yelong Shen}, \bibinfo{person}{Phillip Wallis}, \bibinfo{person}{Zeyuan Allen-Zhu}, \bibinfo{person}{Yuanzhi Li}, \bibinfo{person}{Shean Wang}, \bibinfo{person}{Lu Wang}, {and} \bibinfo{person}{Weizhu Chen}.} \bibinfo{year}{2021}\natexlab{}.
\newblock \showarticletitle{Lora: Low-rank adaptation of large language models}.
\newblock \bibinfo{journal}{\emph{arXiv preprint arXiv:2106.09685}} (\bibinfo{year}{2021}).
\newblock


\bibitem[Huang et~al\mbox{.}(2024)]%
        {huang2024understanding}
\bibfield{author}{\bibinfo{person}{Xu Huang}, \bibinfo{person}{Weiwen Liu}, \bibinfo{person}{Xiaolong Chen}, \bibinfo{person}{Xingmei Wang}, \bibinfo{person}{Hao Wang}, \bibinfo{person}{Defu Lian}, \bibinfo{person}{Yasheng Wang}, \bibinfo{person}{Ruiming Tang}, {and} \bibinfo{person}{Enhong Chen}.} \bibinfo{year}{2024}\natexlab{}.
\newblock \showarticletitle{Understanding the planning of LLM agents: A survey}.
\newblock \bibinfo{journal}{\emph{arXiv preprint arXiv:2402.02716}} (\bibinfo{year}{2024}).
\newblock


\bibitem[J{\"{a}}rvelin and Kek{\"{a}}l{\"{a}}inen(2002)]%
        {DBLP:journals/tois/JarvelinK02}
\bibfield{author}{\bibinfo{person}{Kalervo J{\"{a}}rvelin} {and} \bibinfo{person}{Jaana Kek{\"{a}}l{\"{a}}inen}.} \bibinfo{year}{2002}\natexlab{}.
\newblock \showarticletitle{Cumulated gain-based evaluation of {IR} techniques}.
\newblock \bibinfo{journal}{\emph{{ACM} Trans. Inf. Syst.}} \bibinfo{volume}{20}, \bibinfo{number}{4} (\bibinfo{year}{2002}), \bibinfo{pages}{422--446}.
\newblock
\urldef\tempurl%
\url{https://doi.org/10.1145/582415.582418}
\showDOI{\tempurl}


\bibitem[Jin et~al\mbox{.}(2023)]%
        {jin2023amazon}
\bibfield{author}{\bibinfo{person}{Wei Jin}, \bibinfo{person}{Haitao Mao}, \bibinfo{person}{Zheng Li}, \bibinfo{person}{Haoming Jiang}, \bibinfo{person}{Chen Luo}, \bibinfo{person}{Hongzhi Wen}, \bibinfo{person}{Haoyu Han}, \bibinfo{person}{Hanqing Lu}, \bibinfo{person}{Zhengyang Wang}, \bibinfo{person}{Ruirui Li}, {et~al\mbox{.}}} \bibinfo{year}{2023}\natexlab{}.
\newblock \showarticletitle{Amazon-m2: A multilingual multi-locale shopping session dataset for recommendation and text generation}.
\newblock \bibinfo{journal}{\emph{arXiv preprint arXiv:2307.09688}} (\bibinfo{year}{2023}).
\newblock


\bibitem[Kang and McAuley(2018a)]%
        {DBLP:conf/icdm/KangM18}
\bibfield{author}{\bibinfo{person}{Wang{-}Cheng Kang} {and} \bibinfo{person}{Julian~J. McAuley}.} \bibinfo{year}{2018}\natexlab{a}.
\newblock \showarticletitle{Self-Attentive Sequential Recommendation}. In \bibinfo{booktitle}{\emph{{IEEE} International Conference on Data Mining, {ICDM} 2018, Singapore, November 17-20, 2018}}. \bibinfo{publisher}{{IEEE} Computer Society}, \bibinfo{pages}{197--206}.
\newblock
\urldef\tempurl%
\url{https://doi.org/10.1109/ICDM.2018.00035}
\showDOI{\tempurl}


\bibitem[Kang and McAuley(2018b)]%
        {8594844}
\bibfield{author}{\bibinfo{person}{Wang-Cheng Kang} {and} \bibinfo{person}{Julian McAuley}.} \bibinfo{year}{2018}\natexlab{b}.
\newblock \showarticletitle{Self-Attentive Sequential Recommendation}. In \bibinfo{booktitle}{\emph{2018 IEEE International Conference on Data Mining (ICDM)}}. \bibinfo{pages}{197--206}.
\newblock
\urldef\tempurl%
\url{https://doi.org/10.1109/ICDM.2018.00035}
\showDOI{\tempurl}


\bibitem[Kingma and Ba(2017)]%
        {kingma2017adam}
\bibfield{author}{\bibinfo{person}{Diederik~P. Kingma} {and} \bibinfo{person}{Jimmy Ba}.} \bibinfo{year}{2017}\natexlab{}.
\newblock \bibinfo{title}{Adam: A Method for Stochastic Optimization}.
\newblock
\newblock
\showeprint[arxiv]{1412.6980}~[cs.LG]


\bibitem[Krichene and Rendle(2020)]%
        {krichene2020sampled}
\bibfield{author}{\bibinfo{person}{Walid Krichene} {and} \bibinfo{person}{Steffen Rendle}.} \bibinfo{year}{2020}\natexlab{}.
\newblock \showarticletitle{On sampled metrics for item recommendation}. In \bibinfo{booktitle}{\emph{Proceedings of the 26th ACM SIGKDD international conference on knowledge discovery \& data mining}}. \bibinfo{pages}{1748--1757}.
\newblock


\bibitem[Lei et~al\mbox{.}(2023)]%
        {DBLP:journals/corr/abs-2311-10947}
\bibfield{author}{\bibinfo{person}{Yuxuan Lei}, \bibinfo{person}{Jianxun Lian}, \bibinfo{person}{Jing Yao}, \bibinfo{person}{Xu Huang}, \bibinfo{person}{Defu Lian}, {and} \bibinfo{person}{Xing Xie}.} \bibinfo{year}{2023}\natexlab{}.
\newblock \showarticletitle{RecExplainer: Aligning Large Language Models for Recommendation Model Interpretability}.
\newblock \bibinfo{journal}{\emph{CoRR}}  \bibinfo{volume}{abs/2311.10947} (\bibinfo{year}{2023}).
\newblock
\urldef\tempurl%
\url{https://doi.org/10.48550/ARXIV.2311.10947}
\showDOI{\tempurl}
\showeprint[arXiv]{2311.10947}


\bibitem[Li et~al\mbox{.}(2022)]%
        {10.1145/3488560.3498388}
\bibfield{author}{\bibinfo{person}{Chenglin Li}, \bibinfo{person}{Mingjun Zhao}, \bibinfo{person}{Huanming Zhang}, \bibinfo{person}{Chenyun Yu}, \bibinfo{person}{Lei Cheng}, \bibinfo{person}{Guoqiang Shu}, \bibinfo{person}{BeiBei Kong}, {and} \bibinfo{person}{Di Niu}.} \bibinfo{year}{2022}\natexlab{}.
\newblock \showarticletitle{RecGURU: Adversarial Learning of Generalized User Representations for Cross-Domain Recommendation}. In \bibinfo{booktitle}{\emph{Proceedings of the Fifteenth ACM International Conference on Web Search and Data Mining}} (Virtual Event, AZ, USA) \emph{(\bibinfo{series}{WSDM '22})}. \bibinfo{publisher}{Association for Computing Machinery}, \bibinfo{address}{New York, NY, USA}, \bibinfo{pages}{571–581}.
\newblock
\showISBNx{9781450391320}
\urldef\tempurl%
\url{https://doi.org/10.1145/3488560.3498388}
\showDOI{\tempurl}


\bibitem[Li et~al\mbox{.}(2023c)]%
        {DBLP:conf/ecom/LiZWXLM23}
\bibfield{author}{\bibinfo{person}{Jinming Li}, \bibinfo{person}{Wentao Zhang}, \bibinfo{person}{Tian Wang}, \bibinfo{person}{Guanglei Xiong}, \bibinfo{person}{Alan Lu}, {and} \bibinfo{person}{Gerard Medioni}.} \bibinfo{year}{2023}\natexlab{c}.
\newblock \showarticletitle{GPT4Rec: {A} Generative Framework for Personalized Recommendation and User Interests Interpretation}. In \bibinfo{booktitle}{\emph{Proceedings of the 2023 {SIGIR} Workshop on eCommerce co-located with the 46th International {ACM} {SIGIR} Conference on Research and Development in Information Retrieval {(SIGIR} 2023), Taipei, Taiwan, July 27, 2023}} \emph{(\bibinfo{series}{{CEUR} Workshop Proceedings}, Vol.~\bibinfo{volume}{3589})}, \bibfield{editor}{\bibinfo{person}{Surya Kallumadi}, \bibinfo{person}{Yubin Kim}, \bibinfo{person}{Tracy~Holloway King}, \bibinfo{person}{Shervin Malmasi}, \bibinfo{person}{Maarten de~Rijke}, {and} \bibinfo{person}{Jacopo Tagliabue}} (Eds.). \bibinfo{publisher}{CEUR-WS.org}.
\newblock
\urldef\tempurl%
\url{https://ceur-ws.org/Vol-3589/paper\_2.pdf}
\showURL{%
\tempurl}


\bibitem[Li et~al\mbox{.}(2023b)]%
        {li2023selfprompting}
\bibfield{author}{\bibinfo{person}{Junlong Li}, \bibinfo{person}{Zhuosheng Zhang}, {and} \bibinfo{person}{Hai Zhao}.} \bibinfo{year}{2023}\natexlab{b}.
\newblock \bibinfo{title}{Self-Prompting Large Language Models for Zero-Shot Open-Domain QA}.
\newblock
\newblock
\showeprint[arxiv]{2212.08635}~[cs.CL]


\bibitem[Li et~al\mbox{.}(2023a)]%
        {DBLP:journals/corr/abs-2306-02841}
\bibfield{author}{\bibinfo{person}{Xiangyang Li}, \bibinfo{person}{Bo Chen}, \bibinfo{person}{Lu Hou}, {and} \bibinfo{person}{Ruiming Tang}.} \bibinfo{year}{2023}\natexlab{a}.
\newblock \showarticletitle{{CTRL:} Connect Tabular and Language Model for {CTR} Prediction}.
\newblock \bibinfo{journal}{\emph{CoRR}}  \bibinfo{volume}{abs/2306.02841} (\bibinfo{year}{2023}).
\newblock
\urldef\tempurl%
\url{https://doi.org/10.48550/ARXIV.2306.02841}
\showDOI{\tempurl}
\showeprint[arXiv]{2306.02841}


\bibitem[Liao et~al\mbox{.}(2023)]%
        {DBLP:journals/corr/abs-2312-02445}
\bibfield{author}{\bibinfo{person}{Jiayi Liao}, \bibinfo{person}{Sihang Li}, \bibinfo{person}{Zhengyi Yang}, \bibinfo{person}{Jiancan Wu}, \bibinfo{person}{Yancheng Yuan}, {and} \bibinfo{person}{Xiang Wang}.} \bibinfo{year}{2023}\natexlab{}.
\newblock \showarticletitle{LLaRA: Aligning Large Language Models with Sequential Recommenders}.
\newblock \bibinfo{journal}{\emph{CoRR}}  \bibinfo{volume}{abs/2312.02445} (\bibinfo{year}{2023}).
\newblock
\urldef\tempurl%
\url{https://doi.org/10.48550/ARXIV.2312.02445}
\showDOI{\tempurl}
\showeprint[arXiv]{2312.02445}


\bibitem[Liu et~al\mbox{.}(2023)]%
        {liu2023user}
\bibfield{author}{\bibinfo{person}{Weiwen Liu}, \bibinfo{person}{Wei Guo}, \bibinfo{person}{Yong Liu}, \bibinfo{person}{Ruiming Tang}, {and} \bibinfo{person}{Hao Wang}.} \bibinfo{year}{2023}\natexlab{}.
\newblock \showarticletitle{User Behavior Modeling with Deep Learning for Recommendation: Recent Advances}. In \bibinfo{booktitle}{\emph{Proceedings of the 17th ACM Conference on Recommender Systems}}. \bibinfo{pages}{1286--1287}.
\newblock


\bibitem[Liu et~al\mbox{.}(2021)]%
        {DBLP:journals/corr/abs-2108-06479}
\bibfield{author}{\bibinfo{person}{Zhiwei Liu}, \bibinfo{person}{Yongjun Chen}, \bibinfo{person}{Jia Li}, \bibinfo{person}{Philip~S. Yu}, \bibinfo{person}{Julian~J. McAuley}, {and} \bibinfo{person}{Caiming Xiong}.} \bibinfo{year}{2021}\natexlab{}.
\newblock \showarticletitle{Contrastive Self-supervised Sequential Recommendation with Robust Augmentation}.
\newblock \bibinfo{journal}{\emph{CoRR}}  \bibinfo{volume}{abs/2108.06479} (\bibinfo{year}{2021}).
\newblock
\showeprint[arXiv]{2108.06479}
\urldef\tempurl%
\url{https://arxiv.org/abs/2108.06479}
\showURL{%
\tempurl}


\bibitem[Ma et~al\mbox{.}(2022)]%
        {DBLP:journals/tkdd/MaRCRZLMR22}
\bibfield{author}{\bibinfo{person}{Muyang Ma}, \bibinfo{person}{Pengjie Ren}, \bibinfo{person}{Zhumin Chen}, \bibinfo{person}{Zhaochun Ren}, \bibinfo{person}{Lifan Zhao}, \bibinfo{person}{Peiyu Liu}, \bibinfo{person}{Jun Ma}, {and} \bibinfo{person}{Maarten de Rijke}.} \bibinfo{year}{2022}\natexlab{}.
\newblock \showarticletitle{Mixed Information Flow for Cross-Domain Sequential Recommendations}.
\newblock \bibinfo{journal}{\emph{{ACM} Trans. Knowl. Discov. Data}} \bibinfo{volume}{16}, \bibinfo{number}{4} (\bibinfo{year}{2022}), \bibinfo{pages}{64:1--64:32}.
\newblock
\urldef\tempurl%
\url{https://doi.org/10.1145/3487331}
\showDOI{\tempurl}


\bibitem[Ma et~al\mbox{.}(2019)]%
        {10.1145/3331184.3331200}
\bibfield{author}{\bibinfo{person}{Muyang Ma}, \bibinfo{person}{Pengjie Ren}, \bibinfo{person}{Yujie Lin}, \bibinfo{person}{Zhumin Chen}, \bibinfo{person}{Jun Ma}, {and} \bibinfo{person}{Maarten~de Rijke}.} \bibinfo{year}{2019}\natexlab{}.
\newblock \showarticletitle{$\pi$-Net: A Parallel Information-sharing Network for Shared-account Cross-domain Sequential Recommendations}. In \bibinfo{booktitle}{\emph{Proceedings of the 42nd International ACM SIGIR Conference on Research and Development in Information Retrieval}} (Paris, France) \emph{(\bibinfo{series}{SIGIR'19})}. \bibinfo{publisher}{Association for Computing Machinery}, \bibinfo{address}{New York, NY, USA}, \bibinfo{pages}{685–694}.
\newblock
\showISBNx{9781450361729}
\urldef\tempurl%
\url{https://doi.org/10.1145/3331184.3331200}
\showDOI{\tempurl}


\bibitem[Robertson and Zaragoza(2009)]%
        {10.1561/1500000019}
\bibfield{author}{\bibinfo{person}{Stephen Robertson} {and} \bibinfo{person}{Hugo Zaragoza}.} \bibinfo{year}{2009}\natexlab{}.
\newblock \showarticletitle{The Probabilistic Relevance Framework: BM25 and Beyond}.
\newblock \bibinfo{journal}{\emph{Found. Trends Inf. Retr.}} \bibinfo{volume}{3}, \bibinfo{number}{4} (\bibinfo{date}{apr} \bibinfo{year}{2009}), \bibinfo{pages}{333–389}.
\newblock
\showISSN{1554-0669}
\urldef\tempurl%
\url{https://doi.org/10.1561/1500000019}
\showDOI{\tempurl}


\bibitem[Schein et~al\mbox{.}(2002)]%
        {DBLP:conf/sigir/ScheinPUP02}
\bibfield{author}{\bibinfo{person}{Andrew~I. Schein}, \bibinfo{person}{Alexandrin Popescul}, \bibinfo{person}{Lyle~H. Ungar}, {and} \bibinfo{person}{David~M. Pennock}.} \bibinfo{year}{2002}\natexlab{}.
\newblock \showarticletitle{Methods and metrics for cold-start recommendations}. In \bibinfo{booktitle}{\emph{{SIGIR} 2002: Proceedings of the 25th Annual International {ACM} {SIGIR} Conference on Research and Development in Information Retrieval, August 11-15, 2002, Tampere, Finland}}, \bibfield{editor}{\bibinfo{person}{Kalervo J{\"{a}}rvelin}, \bibinfo{person}{Micheline Beaulieu}, \bibinfo{person}{Ricardo~A. Baeza{-}Yates}, {and} \bibinfo{person}{Sung{-}Hyon Myaeng}} (Eds.). \bibinfo{publisher}{{ACM}}, \bibinfo{pages}{253--260}.
\newblock
\urldef\tempurl%
\url{https://doi.org/10.1145/564376.564421}
\showDOI{\tempurl}


\bibitem[Sun et~al\mbox{.}(2019a)]%
        {10.1145/3357384.3357895}
\bibfield{author}{\bibinfo{person}{Fei Sun}, \bibinfo{person}{Jun Liu}, \bibinfo{person}{Jian Wu}, \bibinfo{person}{Changhua Pei}, \bibinfo{person}{Xiao Lin}, \bibinfo{person}{Wenwu Ou}, {and} \bibinfo{person}{Peng Jiang}.} \bibinfo{year}{2019}\natexlab{a}.
\newblock \showarticletitle{BERT4Rec: Sequential Recommendation with Bidirectional Encoder Representations from Transformer}. In \bibinfo{booktitle}{\emph{Proceedings of the 28th ACM International Conference on Information and Knowledge Management}} (Beijing, China) \emph{(\bibinfo{series}{CIKM '19})}. \bibinfo{publisher}{Association for Computing Machinery}, \bibinfo{address}{New York, NY, USA}, \bibinfo{pages}{1441–1450}.
\newblock
\showISBNx{9781450369763}
\urldef\tempurl%
\url{https://doi.org/10.1145/3357384.3357895}
\showDOI{\tempurl}


\bibitem[Sun et~al\mbox{.}(2019b)]%
        {DBLP:conf/cikm/SunLWPLOJ19}
\bibfield{author}{\bibinfo{person}{Fei Sun}, \bibinfo{person}{Jun Liu}, \bibinfo{person}{Jian Wu}, \bibinfo{person}{Changhua Pei}, \bibinfo{person}{Xiao Lin}, \bibinfo{person}{Wenwu Ou}, {and} \bibinfo{person}{Peng Jiang}.} \bibinfo{year}{2019}\natexlab{b}.
\newblock \showarticletitle{BERT4Rec: Sequential Recommendation with Bidirectional Encoder Representations from Transformer}. In \bibinfo{booktitle}{\emph{Proceedings of the 28th {ACM} International Conference on Information and Knowledge Management, {CIKM} 2019, Beijing, China, November 3-7, 2019}}, \bibfield{editor}{\bibinfo{person}{Wenwu Zhu}, \bibinfo{person}{Dacheng Tao}, \bibinfo{person}{Xueqi Cheng}, \bibinfo{person}{Peng Cui}, \bibinfo{person}{Elke~A. Rundensteiner}, \bibinfo{person}{David Carmel}, \bibinfo{person}{Qi~He}, {and} \bibinfo{person}{Jeffrey~Xu Yu}} (Eds.). \bibinfo{publisher}{{ACM}}, \bibinfo{pages}{1441--1450}.
\newblock
\urldef\tempurl%
\url{https://doi.org/10.1145/3357384.3357895}
\showDOI{\tempurl}


\bibitem[Sun et~al\mbox{.}(2023a)]%
        {9647967}
\bibfield{author}{\bibinfo{person}{Wenchao Sun}, \bibinfo{person}{Muyang Ma}, \bibinfo{person}{Pengjie Ren}, \bibinfo{person}{Yujie Lin}, \bibinfo{person}{Zhumin Chen}, \bibinfo{person}{Zhaochun Ren}, \bibinfo{person}{Jun Ma}, {and} \bibinfo{person}{Maarten de Rijke}.} \bibinfo{year}{2023}\natexlab{a}.
\newblock \showarticletitle{Parallel Split-Join Networks for Shared Account Cross-Domain Sequential Recommendations}.
\newblock \bibinfo{journal}{\emph{IEEE Transactions on Knowledge and Data Engineering}} \bibinfo{volume}{35}, \bibinfo{number}{4} (\bibinfo{year}{2023}), \bibinfo{pages}{4106--4123}.
\newblock
\urldef\tempurl%
\url{https://doi.org/10.1109/TKDE.2021.3130927}
\showDOI{\tempurl}


\bibitem[Sun et~al\mbox{.}(2023b)]%
        {sun-etal-2023-chatgpt}
\bibfield{author}{\bibinfo{person}{Weiwei Sun}, \bibinfo{person}{Lingyong Yan}, \bibinfo{person}{Xinyu Ma}, \bibinfo{person}{Shuaiqiang Wang}, \bibinfo{person}{Pengjie Ren}, \bibinfo{person}{Zhumin Chen}, \bibinfo{person}{Dawei Yin}, {and} \bibinfo{person}{Zhaochun Ren}.} \bibinfo{year}{2023}\natexlab{b}.
\newblock \showarticletitle{Is {C}hat{GPT} Good at Search? Investigating Large Language Models as Re-Ranking Agents}. In \bibinfo{booktitle}{\emph{Proceedings of the 2023 Conference on Empirical Methods in Natural Language Processing}}, \bibfield{editor}{\bibinfo{person}{Houda Bouamor}, \bibinfo{person}{Juan Pino}, {and} \bibinfo{person}{Kalika Bali}} (Eds.). \bibinfo{publisher}{Association for Computational Linguistics}, \bibinfo{address}{Singapore}, \bibinfo{pages}{14918--14937}.
\newblock
\urldef\tempurl%
\url{https://doi.org/10.18653/v1/2023.emnlp-main.923}
\showDOI{\tempurl}


\bibitem[Tang and Wang(2018)]%
        {10.1145/3159652.3159656}
\bibfield{author}{\bibinfo{person}{Jiaxi Tang} {and} \bibinfo{person}{Ke Wang}.} \bibinfo{year}{2018}\natexlab{}.
\newblock \showarticletitle{Personalized Top-N Sequential Recommendation via Convolutional Sequence Embedding}. In \bibinfo{booktitle}{\emph{Proceedings of the Eleventh ACM International Conference on Web Search and Data Mining}} (Marina Del Rey, CA, USA) \emph{(\bibinfo{series}{WSDM '18})}. \bibinfo{publisher}{Association for Computing Machinery}, \bibinfo{address}{New York, NY, USA}, \bibinfo{pages}{565–573}.
\newblock
\showISBNx{9781450355810}
\urldef\tempurl%
\url{https://doi.org/10.1145/3159652.3159656}
\showDOI{\tempurl}


\bibitem[Tang et~al\mbox{.}(2023)]%
        {DBLP:journals/corr/abs-2310-14304}
\bibfield{author}{\bibinfo{person}{Zuoli Tang}, \bibinfo{person}{Zhaoxin Huan}, \bibinfo{person}{Zihao Li}, \bibinfo{person}{Xiaolu Zhang}, \bibinfo{person}{Jun Hu}, \bibinfo{person}{Chilin Fu}, \bibinfo{person}{Jun Zhou}, {and} \bibinfo{person}{Chenliang Li}.} \bibinfo{year}{2023}\natexlab{}.
\newblock \showarticletitle{One Model for All: Large Language Models are Domain-Agnostic Recommendation Systems}.
\newblock \bibinfo{journal}{\emph{CoRR}}  \bibinfo{volume}{abs/2310.14304} (\bibinfo{year}{2023}).
\newblock
\urldef\tempurl%
\url{https://doi.org/10.48550/ARXIV.2310.14304}
\showDOI{\tempurl}
\showeprint[arXiv]{2310.14304}


\bibitem[Voorhees(1999)]%
        {DBLP:conf/trec/Voorhees99}
\bibfield{author}{\bibinfo{person}{Ellen~M. Voorhees}.} \bibinfo{year}{1999}\natexlab{}.
\newblock \showarticletitle{The {TREC-8} Question Answering Track Report}. In \bibinfo{booktitle}{\emph{Proceedings of The Eighth Text REtrieval Conference, {TREC} 1999, Gaithersburg, Maryland, USA, November 17-19, 1999}} \emph{(\bibinfo{series}{{NIST} Special Publication}, Vol.~\bibinfo{volume}{500-246})}, \bibfield{editor}{\bibinfo{person}{Ellen~M. Voorhees} {and} \bibinfo{person}{Donna~K. Harman}} (Eds.). \bibinfo{publisher}{National Institute of Standards and Technology {(NIST)}}.
\newblock
\urldef\tempurl%
\url{http://trec.nist.gov/pubs/trec8/papers/qa\_report.pdf}
\showURL{%
\tempurl}


\bibitem[Wang et~al\mbox{.}(2021)]%
        {wang2021hypersorec}
\bibfield{author}{\bibinfo{person}{Hao Wang}, \bibinfo{person}{Defu Lian}, \bibinfo{person}{Hanghang Tong}, \bibinfo{person}{Qi Liu}, \bibinfo{person}{Zhenya Huang}, {and} \bibinfo{person}{Enhong Chen}.} \bibinfo{year}{2021}\natexlab{}.
\newblock \showarticletitle{Hypersorec: Exploiting hyperbolic user and item representations with multiple aspects for social-aware recommendation}.
\newblock \bibinfo{journal}{\emph{ACM Transactions on Information Systems (TOIS)}} \bibinfo{volume}{40}, \bibinfo{number}{2} (\bibinfo{year}{2021}), \bibinfo{pages}{1--28}.
\newblock


\bibitem[Wang et~al\mbox{.}(2019)]%
        {wang2019mcne}
\bibfield{author}{\bibinfo{person}{Hao Wang}, \bibinfo{person}{Tong Xu}, \bibinfo{person}{Qi Liu}, \bibinfo{person}{Defu Lian}, \bibinfo{person}{Enhong Chen}, \bibinfo{person}{Dongfang Du}, \bibinfo{person}{Han Wu}, {and} \bibinfo{person}{Wen Su}.} \bibinfo{year}{2019}\natexlab{}.
\newblock \showarticletitle{MCNE: An end-to-end framework for learning multiple conditional network representations of social network}. In \bibinfo{booktitle}{\emph{Proceedings of the 25th ACM SIGKDD international conference on knowledge discovery \& data mining}}. \bibinfo{pages}{1064--1072}.
\newblock


\bibitem[Wang et~al\mbox{.}(2023)]%
        {DBLP:journals/corr/abs-2308-14296}
\bibfield{author}{\bibinfo{person}{Yancheng Wang}, \bibinfo{person}{Ziyan Jiang}, \bibinfo{person}{Zheng Chen}, \bibinfo{person}{Fan Yang}, \bibinfo{person}{Yingxue Zhou}, \bibinfo{person}{Eunah Cho}, \bibinfo{person}{Xing Fan}, \bibinfo{person}{Xiaojiang Huang}, \bibinfo{person}{Yanbin Lu}, {and} \bibinfo{person}{Yingzhen Yang}.} \bibinfo{year}{2023}\natexlab{}.
\newblock \showarticletitle{RecMind: Large Language Model Powered Agent For Recommendation}.
\newblock \bibinfo{journal}{\emph{CoRR}}  \bibinfo{volume}{abs/2308.14296} (\bibinfo{year}{2023}).
\newblock
\urldef\tempurl%
\url{https://doi.org/10.48550/ARXIV.2308.14296}
\showDOI{\tempurl}
\showeprint[arXiv]{2308.14296}


\bibitem[Wang et~al\mbox{.}(2020)]%
        {10.1145/3397271.3401142}
\bibfield{author}{\bibinfo{person}{Ziyang Wang}, \bibinfo{person}{Wei Wei}, \bibinfo{person}{Gao Cong}, \bibinfo{person}{Xiao-Li Li}, \bibinfo{person}{Xian-Ling Mao}, {and} \bibinfo{person}{Minghui Qiu}.} \bibinfo{year}{2020}\natexlab{}.
\newblock \showarticletitle{Global Context Enhanced Graph Neural Networks for Session-based Recommendation}. In \bibinfo{booktitle}{\emph{Proceedings of the 43rd International ACM SIGIR Conference on Research and Development in Information Retrieval}} (Virtual Event, China) \emph{(\bibinfo{series}{SIGIR '20})}. \bibinfo{publisher}{Association for Computing Machinery}, \bibinfo{address}{New York, NY, USA}, \bibinfo{pages}{169–178}.
\newblock
\showISBNx{9781450380164}
\urldef\tempurl%
\url{https://doi.org/10.1145/3397271.3401142}
\showDOI{\tempurl}


\bibitem[Waters(1976)]%
        {waters1976hit}
\bibfield{author}{\bibinfo{person}{SJ Waters}.} \bibinfo{year}{1976}\natexlab{}.
\newblock \showarticletitle{Hit ratios}.
\newblock \bibinfo{journal}{\emph{Comput. J.}} \bibinfo{volume}{19}, \bibinfo{number}{1} (\bibinfo{year}{1976}), \bibinfo{pages}{21--24}.
\newblock


\bibitem[Wei et~al\mbox{.}(2022)]%
        {DBLP:conf/nips/Wei0SBIXCLZ22}
\bibfield{author}{\bibinfo{person}{Jason Wei}, \bibinfo{person}{Xuezhi Wang}, \bibinfo{person}{Dale Schuurmans}, \bibinfo{person}{Maarten Bosma}, \bibinfo{person}{Brian Ichter}, \bibinfo{person}{Fei Xia}, \bibinfo{person}{Ed~H. Chi}, \bibinfo{person}{Quoc~V. Le}, {and} \bibinfo{person}{Denny Zhou}.} \bibinfo{year}{2022}\natexlab{}.
\newblock \showarticletitle{Chain-of-Thought Prompting Elicits Reasoning in Large Language Models}. In \bibinfo{booktitle}{\emph{Advances in Neural Information Processing Systems 35: Annual Conference on Neural Information Processing Systems 2022, NeurIPS 2022, New Orleans, LA, USA, November 28 - December 9, 2022}}, \bibfield{editor}{\bibinfo{person}{Sanmi Koyejo}, \bibinfo{person}{S.~Mohamed}, \bibinfo{person}{A.~Agarwal}, \bibinfo{person}{Danielle Belgrave}, \bibinfo{person}{K.~Cho}, {and} \bibinfo{person}{A.~Oh}} (Eds.).
\newblock
\urldef\tempurl%
\url{http://papers.nips.cc/paper\_files/paper/2022/hash/9d5609613524ecf4f15af0f7b31abca4-Abstract-Conference.html}
\showURL{%
\tempurl}


\bibitem[Wu et~al\mbox{.}(2023)]%
        {wu2023survey}
\bibfield{author}{\bibinfo{person}{Likang Wu}, \bibinfo{person}{Zhi Zheng}, \bibinfo{person}{Zhaopeng Qiu}, \bibinfo{person}{Hao Wang}, \bibinfo{person}{Hongchao Gu}, \bibinfo{person}{Tingjia Shen}, \bibinfo{person}{Chuan Qin}, \bibinfo{person}{Chen Zhu}, \bibinfo{person}{Hengshu Zhu}, \bibinfo{person}{Qi Liu}, {et~al\mbox{.}}} \bibinfo{year}{2023}\natexlab{}.
\newblock \showarticletitle{A Survey on Large Language Models for Recommendation}.
\newblock \bibinfo{journal}{\emph{arXiv preprint arXiv:2305.19860}} (\bibinfo{year}{2023}).
\newblock


\bibitem[Xie et~al\mbox{.}(2022)]%
        {9835621}
\bibfield{author}{\bibinfo{person}{Xu Xie}, \bibinfo{person}{Fei Sun}, \bibinfo{person}{Zhaoyang Liu}, \bibinfo{person}{Shiwen Wu}, \bibinfo{person}{Jinyang Gao}, \bibinfo{person}{Jiandong Zhang}, \bibinfo{person}{Bolin Ding}, {and} \bibinfo{person}{Bin Cui}.} \bibinfo{year}{2022}\natexlab{}.
\newblock \showarticletitle{Contrastive Learning for Sequential Recommendation}. In \bibinfo{booktitle}{\emph{2022 IEEE 38th International Conference on Data Engineering (ICDE)}}. \bibinfo{pages}{1259--1273}.
\newblock
\urldef\tempurl%
\url{https://doi.org/10.1109/ICDE53745.2022.00099}
\showDOI{\tempurl}


\bibitem[Yang et~al\mbox{.}(2023)]%
        {DBLP:journals/corr/abs-2310-20487}
\bibfield{author}{\bibinfo{person}{Zhengyi Yang}, \bibinfo{person}{Jiancan Wu}, \bibinfo{person}{Yanchen Luo}, \bibinfo{person}{Jizhi Zhang}, \bibinfo{person}{Yancheng Yuan}, \bibinfo{person}{An Zhang}, \bibinfo{person}{Xiang Wang}, {and} \bibinfo{person}{Xiangnan He}.} \bibinfo{year}{2023}\natexlab{}.
\newblock \showarticletitle{Large Language Model Can Interpret Latent Space of Sequential Recommender}.
\newblock \bibinfo{journal}{\emph{CoRR}}  \bibinfo{volume}{abs/2310.20487} (\bibinfo{year}{2023}).
\newblock
\urldef\tempurl%
\url{https://doi.org/10.48550/ARXIV.2310.20487}
\showDOI{\tempurl}
\showeprint[arXiv]{2310.20487}


\bibitem[Yin et~al\mbox{.}(2024a)]%
        {yin2024learning}
\bibfield{author}{\bibinfo{person}{Mingjia Yin}, \bibinfo{person}{Hao Wang}, \bibinfo{person}{Wei Guo}, \bibinfo{person}{Yong Liu}, \bibinfo{person}{Zhi Li}, \bibinfo{person}{Sirui Zhao}, \bibinfo{person}{Defu Lian}, {and} \bibinfo{person}{Enhong Chen}.} \bibinfo{year}{2024}\natexlab{a}.
\newblock \showarticletitle{Learning Partially Aligned Item Representation for Cross-Domain Sequential Recommendation}.
\newblock \bibinfo{journal}{\emph{arXiv preprint arXiv:2405.12473}} (\bibinfo{year}{2024}).
\newblock


\bibitem[Yin et~al\mbox{.}(2024b)]%
        {yin2024dataset}
\bibfield{author}{\bibinfo{person}{Mingjia Yin}, \bibinfo{person}{Hao Wang}, \bibinfo{person}{Wei Guo}, \bibinfo{person}{Yong Liu}, \bibinfo{person}{Suojuan Zhang}, \bibinfo{person}{Sirui Zhao}, \bibinfo{person}{Defu Lian}, {and} \bibinfo{person}{Enhong Chen}.} \bibinfo{year}{2024}\natexlab{b}.
\newblock \showarticletitle{Dataset Regeneration for Sequential Recommendation}.
\newblock \bibinfo{journal}{\emph{arXiv preprint arXiv:2405.17795}} (\bibinfo{year}{2024}).
\newblock


\bibitem[Yin et~al\mbox{.}(2023)]%
        {yin2023apgl4sr}
\bibfield{author}{\bibinfo{person}{Mingjia Yin}, \bibinfo{person}{Hao Wang}, \bibinfo{person}{Xiang Xu}, \bibinfo{person}{Likang Wu}, \bibinfo{person}{Sirui Zhao}, \bibinfo{person}{Wei Guo}, \bibinfo{person}{Yong Liu}, \bibinfo{person}{Ruiming Tang}, \bibinfo{person}{Defu Lian}, {and} \bibinfo{person}{Enhong Chen}.} \bibinfo{year}{2023}\natexlab{}.
\newblock \showarticletitle{APGL4SR: A Generic Framework with Adaptive and Personalized Global Collaborative Information in Sequential Recommendation}. In \bibinfo{booktitle}{\emph{Proceedings of the 32nd ACM International Conference on Information and Knowledge Management}}. \bibinfo{pages}{3009--3019}.
\newblock


\bibitem[Yuan et~al\mbox{.}(2023)]%
        {DBLP:conf/sigir/YuanYSLFYPN23}
\bibfield{author}{\bibinfo{person}{Zheng Yuan}, \bibinfo{person}{Fajie Yuan}, \bibinfo{person}{Yu Song}, \bibinfo{person}{Youhua Li}, \bibinfo{person}{Junchen Fu}, \bibinfo{person}{Fei Yang}, \bibinfo{person}{Yunzhu Pan}, {and} \bibinfo{person}{Yongxin Ni}.} \bibinfo{year}{2023}\natexlab{}.
\newblock \showarticletitle{Where to Go Next for Recommender Systems? {ID-} vs. Modality-based Recommender Models Revisited}. In \bibinfo{booktitle}{\emph{Proceedings of the 46th International {ACM} {SIGIR} Conference on Research and Development in Information Retrieval, {SIGIR} 2023, Taipei, Taiwan, July 23-27, 2023}}, \bibfield{editor}{\bibinfo{person}{Hsin{-}Hsi Chen}, \bibinfo{person}{Wei{-}Jou~(Edward) Duh}, \bibinfo{person}{Hen{-}Hsen Huang}, \bibinfo{person}{Makoto~P. Kato}, \bibinfo{person}{Josiane Mothe}, {and} \bibinfo{person}{Barbara Poblete}} (Eds.). \bibinfo{publisher}{{ACM}}, \bibinfo{pages}{2639--2649}.
\newblock
\urldef\tempurl%
\url{https://doi.org/10.1145/3539618.3591932}
\showDOI{\tempurl}


\bibitem[Zhang et~al\mbox{.}(2023b)]%
        {zhang2023generative}
\bibfield{author}{\bibinfo{person}{An Zhang}, \bibinfo{person}{Leheng Sheng}, \bibinfo{person}{Yuxin Chen}, \bibinfo{person}{Hao Li}, \bibinfo{person}{Yang Deng}, \bibinfo{person}{Xiang Wang}, {and} \bibinfo{person}{Tat-Seng Chua}.} \bibinfo{year}{2023}\natexlab{b}.
\newblock \bibinfo{title}{On Generative Agents in Recommendation}.
\newblock
\newblock
\showeprint[arxiv]{2310.10108}~[cs.IR]


\bibitem[Zhang et~al\mbox{.}(2024)]%
        {zhang2024unified}
\bibfield{author}{\bibinfo{person}{Luankang Zhang}, \bibinfo{person}{Hao Wang}, \bibinfo{person}{Suojuan Zhang}, \bibinfo{person}{Mingjia Yin}, \bibinfo{person}{Yongqiang Han}, \bibinfo{person}{Jiaqing Zhang}, \bibinfo{person}{Defu Lian}, {and} \bibinfo{person}{Enhong Chen}.} \bibinfo{year}{2024}\natexlab{}.
\newblock \showarticletitle{A Unified Framework for Adaptive Representation Enhancement and Inversed Learning in Cross-Domain Recommendation}.
\newblock \bibinfo{journal}{\emph{arXiv preprint arXiv:2404.00268}} (\bibinfo{year}{2024}).
\newblock


\bibitem[Zhang et~al\mbox{.}(2022)]%
        {zhang2022clustering}
\bibfield{author}{\bibinfo{person}{Yuren Zhang}, \bibinfo{person}{Enhong Chen}, \bibinfo{person}{Binbin Jin}, \bibinfo{person}{Hao Wang}, \bibinfo{person}{Min Hou}, \bibinfo{person}{Wei Huang}, {and} \bibinfo{person}{Runlong Yu}.} \bibinfo{year}{2022}\natexlab{}.
\newblock \showarticletitle{Clustering based behavior sampling with long sequential data for CTR prediction}. In \bibinfo{booktitle}{\emph{Proceedings of the 45th International ACM SIGIR Conference on Research and Development in Information Retrieval}}. \bibinfo{pages}{2195--2200}.
\newblock


\bibitem[Zhang et~al\mbox{.}(2023a)]%
        {DBLP:journals/corr/abs-2310-19488}
\bibfield{author}{\bibinfo{person}{Yang Zhang}, \bibinfo{person}{Fuli Feng}, \bibinfo{person}{Jizhi Zhang}, \bibinfo{person}{Keqin Bao}, \bibinfo{person}{Qifan Wang}, {and} \bibinfo{person}{Xiangnan He}.} \bibinfo{year}{2023}\natexlab{a}.
\newblock \showarticletitle{CoLLM: Integrating Collaborative Embeddings into Large Language Models for Recommendation}.
\newblock \bibinfo{journal}{\emph{CoRR}}  \bibinfo{volume}{abs/2310.19488} (\bibinfo{year}{2023}).
\newblock
\urldef\tempurl%
\url{https://doi.org/10.48550/ARXIV.2310.19488}
\showDOI{\tempurl}
\showeprint[arXiv]{2310.19488}


\bibitem[Zhou et~al\mbox{.}(2020)]%
        {DBLP:conf/cikm/ZhouWZZWZWW20}
\bibfield{author}{\bibinfo{person}{Kun Zhou}, \bibinfo{person}{Hui Wang}, \bibinfo{person}{Wayne~Xin Zhao}, \bibinfo{person}{Yutao Zhu}, \bibinfo{person}{Sirui Wang}, \bibinfo{person}{Fuzheng Zhang}, \bibinfo{person}{Zhongyuan Wang}, {and} \bibinfo{person}{Ji{-}Rong Wen}.} \bibinfo{year}{2020}\natexlab{}.
\newblock \showarticletitle{S3-Rec: Self-Supervised Learning for Sequential Recommendation with Mutual Information Maximization}. In \bibinfo{booktitle}{\emph{{CIKM} '20: The 29th {ACM} International Conference on Information and Knowledge Management, Virtual Event, Ireland, October 19-23, 2020}}, \bibfield{editor}{\bibinfo{person}{Mathieu d'Aquin}, \bibinfo{person}{Stefan Dietze}, \bibinfo{person}{Claudia Hauff}, \bibinfo{person}{Edward Curry}, {and} \bibinfo{person}{Philippe Cudr{\'{e}}{-}Mauroux}} (Eds.). \bibinfo{publisher}{{ACM}}, \bibinfo{pages}{1893--1902}.
\newblock
\urldef\tempurl%
\url{https://doi.org/10.1145/3340531.3411954}
\showDOI{\tempurl}


\end{thebibliography}
\end{document}